\pdfoutput=1
\documentclass{article}

\PassOptionsToPackage{numbers, compress, sort}{natbib}
\usepackage[main, final]{neurips_2026}

\usepackage[utf8]{inputenc}
\usepackage[T1]{fontenc}

\usepackage{url}
\usepackage{booktabs}
\usepackage{amsfonts}
\usepackage{nicefrac}
\usepackage{microtype}
\usepackage{xcolor}
\usepackage{amsmath,amsthm,amssymb,mathtools}
\usepackage{enumitem}
\usepackage{multirow}
\usepackage{graphicx}
\usepackage{tikz}

\IfFileExists{algorithm.sty}{%
  \usepackage{algorithm,algorithmic}%
}{%
  \usepackage{float}%
  \floatstyle{ruled}%
  \newfloat{algorithm}{htbp}{loa}%
  \floatname{algorithm}{Algorithm}%
  \newcounter{algline}%
  \newdimen\algindent%
  \newenvironment{algorithmic}[1][1]{%
    \setcounter{algline}{0}%
    \algindent=0pt%
    \begin{list}{\footnotesize\arabic{algline}:}{%
      \usecounter{algline}%
      \setlength{\leftmargin}{1.8em}%
      \setlength{\itemsep}{1.5pt}%
      \setlength{\parsep}{0pt}%
      \setlength{\topsep}{2pt}%
    }%
  }{\end{list}}%
  \newcommand{\REQUIRE}{\item[\textbf{Require:}]}%
  \newcommand{\STATE}{\item\hangindent=\dimexpr\leftmargin+\algindent\relax\hangafter=1\hspace*{\algindent}}%
  \newcommand{\WHILE}[1]{\item\hangindent=\dimexpr\leftmargin+\algindent\relax\hangafter=1\hspace*{\algindent}\textbf{while} ##1 \textbf{do}\addtolength{\algindent}{1.1em}}%
  \newcommand{\ENDWHILE}{\addtolength{\algindent}{-1.1em}\item\hspace*{\algindent}\textbf{end while}}%
  \newcommand{\FOR}[1]{\item\hangindent=\dimexpr\leftmargin+\algindent\relax\hangafter=1\hspace*{\algindent}\textbf{for} ##1 \textbf{do}\addtolength{\algindent}{1.1em}}%
  \newcommand{\ENDFOR}{\addtolength{\algindent}{-1.1em}\item\hspace*{\algindent}\textbf{end for}}%
}

\usepackage{hyperref}
\hypersetup{pdftitle={MaxIM: Maximally Informative Incremental Summarization via Reinforcement Learning}}

\newtheorem{theorem}{Theorem}
\newtheorem{lemma}{Lemma}
\newtheorem{proposition}{Proposition}
\newtheorem{corollary}{Corollary}
\theoremstyle{remark}

\newcommand{\E}{\mathbb{E}}
\newcommand{\KL}{D_{\mathrm{KL}}}
\newcommand{\TV}{D_{\mathrm{TV}}}
\newcommand{\mcell}[1]{\begin{tabular}[c]{@{}c@{}}#1\end{tabular}}
\newcommand{\lcell}[1]{\begin{tabular}[c]{@{}l@{}}#1\end{tabular}}
\newcommand{\pvar}[1]{\texttt{\detokenize{{#1}}}}
\IfFileExists{tcolorbox.sty}{%
  \usepackage[most]{tcolorbox}%
  \newtcolorbox{promptbox}[1]{%
    enhanced,
    colback=black!2,
    colframe=black!45,
    colbacktitle=black!9,
    coltitle=black,
    fonttitle=\bfseries\footnotesize,
    fontupper=\footnotesize,
    boxrule=0.45pt,
    arc=2pt,
    left=6pt, right=6pt, top=4pt, bottom=4pt,
    toptitle=2.5pt, bottomtitle=2.5pt,
    before skip=5pt, after skip=6pt,
    title={##1}%
  }%
}{%
  \newenvironment{promptbox}[1]{%
    \par\medskip\noindent\fbox{\begin{minipage}{0.97\textwidth}\textbf{\footnotesize ##1}\par\smallskip\footnotesize%
  }{\end{minipage}}\par\medskip}%
}

\title{MaxIM: Maximally Informative Incremental Summarization via Reinforcement Learning}

\author{%
  Jihwan Jeong$^{1}$\hspace{0.75em}Guy Tennenholtz$^{2}$\thanks{Work done while at Google.}\hspace{0.75em}Yinlam Chow$^{3}$\hspace{0.75em}Chih-Wei Hsu$^{1}$\hspace{0.75em}Craig Boutilier$^{1}$\\[3pt]
  $^{1}$Google Research \qquad $^{2}$Noetive \qquad $^{3}$Google DeepMind\\
  {\small\texttt{\{jihwanjeong, yinlamchow, cwhsu, cboutilier\}@google.com}, \texttt{guytenn@gmail.com}}
}

\begin{document}

\maketitle

\begin{abstract}
AI systems must often maintain bounded memory of unbounded interaction histories. However, optimizing the memory representation, and its updates, for downstream task utility is difficult for standard reinforcement learning (RL) due to severe credit assignment bottlenecks induced by sparse, delayed rewards. We introduce \emph{MaxIM (Maximally Informative Incremental Memory)}, a framework that formulates incremental summarization as a capacity-constrained, information-theoretic reward-shaping problem. Grounded in value suboptimality bounds, we show that bounding sequential information loss motivates maximizing the predictive log-likelihood of future, task-relevant targets, a variational lower bound on \emph{predictive sufficiency}. We introduce a \emph{recursive consistency} condition to ensure summaries are grounded by maximizing mutual information (MI) with the observed history. We mitigate credit assignment issues with a parameterized asymmetric critic that computes step-wise variational MI bounds for dense reward shaping. Empirical evaluation on long-horizon scientific reading and multi-turn dialogue benchmarks shows that MaxIM outperforms RL, supervised fine-tuning, and prompt-focused baselines on factual grounding, comprehensive autorater metrics, and downstream task utility across strict token budgets and extended horizons, without the need for expensive human preference labels.
\end{abstract}

\section{Introduction}
\label{sec:intro}

A critical challenge facing AI agents in continuous, streaming environments---from real-time personalization~\cite{covington2016youtube, bao2024realtime} to long-horizon dialogue~\cite{xu2022beyond, xu2022longtime}---is the inability to retain unbounded context. Despite the expanded context windows of modern large language models (LLMs), processing ever-growing interaction logs 
frequently
degrades reasoning performance due to ``lost in the middle'' phenomena~\cite{liu2023lost}. While heuristic memory buffers (e.g., MemGPT~\cite{packer2023memgpt}) and RL-based memory models (e.g., MemAgent~\cite{yu2025memagent}, Mem-$\alpha$~\cite{wang2025memalpha}) offer practical mechanisms for context management, agents ultimately need rigorous methods to actively compress these streams into a fixed-capacity working memory.

We assume a trainable history/context summarization policy manages this bounded memory state, while a separate, frozen action agent uses it to choose how to act in the environment. In this setting, the objective of a memory system is to maximize downstream task utility, $R_{\text{task}}$. In principle, optimizing a summarizer using reinforcement learning (RL) on $R_{\text{task}}$ yields an optimal compression strategy. However, in long-horizon settings, task rewards are notoriously sparse and delayed, inducing extreme credit assignment failures in RL---the policy receives insufficiently granular feedback to learn which observation-stream features are predictive of task success. Conversely, optimizing for surface-level text reconstruction~\cite{chang2024booookscore, hwang2024enhancing} discards deeper latent intents needed for effective future inference.

\begin{figure}[t]
    \centering
    \includegraphics[width=0.80\textwidth]{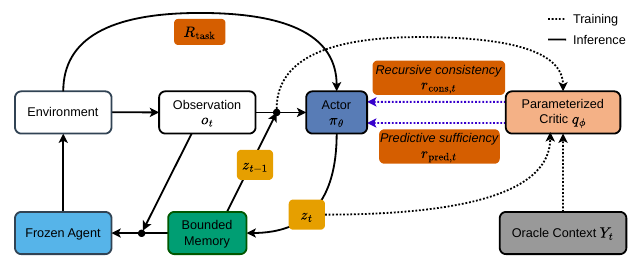}
    \vspace{-2pt}
    \caption{The MaxIM framework. \textbf{Streaming Inference (Solid Lines):} A trained actor policy incrementally compresses prior summaries $z_{t-1}$ and incoming observations $o_t$ into a bounded memory $z_t$. This memory, alongside recent observations, provides the context for a frozen downstream agent interacting with the environment with zero critic calls at test time. \textbf{Asymmetric Training (Dotted Lines):} During training, a pretrained critic $q_\phi$ evaluates the actor's summary using oracle context $Y_t$ and the immediate history $(z_{t-1}, o_t)$. It computes dense, information-theoretic shaping rewards (predictive sufficiency and recursive consistency) that step-wise guide the policy, bypassing the credit assignment bottleneck of the sparse terminal task reward $R_{\text{task}}$.}
    \label{fig:framework}
    \vspace{-6pt}
\end{figure}

To bridge this gap, we develop \emph{MaxIM (Maximally Informative Incremental Memory)}, a novel RL framework that addresses credit assignment with dense information-theoretic reward shaping. Leveraging bounds from state abstraction theory, we show that the performance gap induced by constrained memory is bounded by sequential information loss. We show that bounding this loss motivates maximizing the predictive log-likelihood of future, task-relevant target variables $Y_t$, a variational lower bound on \emph{predictive sufficiency}. However, optimizing an LLM purely for future prediction can induce severe hallucinations (i.e., fabricating plausible-sounding text that aligns with the task/target, but is disconnected from actual observations). To prevent this, we derive a \emph{recursive consistency} regularizer that maximizes the mutual information (MI) with the immediate history, substantially suppressing ungrounded generation.

To optimize this dual-objective, MaxIM employs RL with asymmetric context (see Figure~\ref{fig:framework}). We leverage synthetic datasets to train a parameterized critic using supervised cross-entropy minimization, which computes step-wise baseline-normalized variational predictive likelihood (VPL) bounds~\cite{barber2003vpl}, thus providing dense shaping rewards to the policy. We update the policy using standard policy gradient methods equipped with a backward-discounted reward-to-go to account for temporal error accumulation, while gradually annealing the shaping weight to ensure that late-stage training directly optimizes downstream task utility.

Our contributions are: (1) We develop MaxIM, a principled RL framework for capacity-constrained generative memory that solves the sparse credit assignment bottleneck via dense information-theoretic reward shaping. (2) We link value suboptimality bounds to sequential information loss across both static and dynamic partially observable environments, deriving two complementary regularizers: \emph{predictive sufficiency} and \emph{recursive consistency}. (3) We offer an effective training architecture using scalable synthetic data generation and a unified asymmetric critic to compute calibrated variational lower bounds, obviating the need for human preference labels. (4) We demonstrate MaxIM's efficacy on diverse benchmarks---a dynamic multi-turn user simulator and a novel arXiv scientific reading task---achieving superior factual grounding, token-budget Pareto efficiency, and task success.

\section{Related Work}
\label{sec:related}

\paragraph{Incremental Summarization and Memory Buffers.}
Traditional summarization processes entire documents to generate static outputs~\cite{see2017get, lewis2020bart}, whereas incremental summarization recursively updates a summary as streaming information arrives~\cite{kedzie2016real}. While early text streaming systems relied on clustering algorithms~\cite{allan1998line}, modern neural approaches exploit structural representations (e.g., JSON states) for incremental updates~\cite{hwang2024enhancing}. A parallel line of work compresses long contexts in continuous space---including learned soft-prompt or embedding compression~\cite{mu2023gisting, chevalier2023autocompressor, ge2024icae} and optimal-transport token-merging or KV-cache compression. These methods require either retraining the downstream model to consume continuous latents or white-box access to internal model activations. By contrast, we maintain a discrete, human-readable \emph{text memory} under a token budget $K$, allowing it to interface directly with any frozen black-box agent. Recently, sophisticated memory buffer methods have emerged to manage unbounded LLM contexts. The closest analogs to our setup are memory systems such as MemGPT~\cite{packer2023memgpt}, Mem-$\alpha$~\cite{wang2025memalpha}, and MemAgent~\cite{yu2025memagent}. While MemGPT and Mem-$\alpha$ focus on tool-calling, MemAgent learns an explicit, generative memory update policy using RL, as we do. However, these methods rely on sparse task rewards which can exacerbate hallucination and historical drift due to credit assignment failures. MaxIM builds on this foundation by introducing theoretically derived, dense MI regularizers that explicitly guide the retention of task-relevant information.

\paragraph{Preference Modeling and Information-Theoretic Shaping.}
Modern alignment techniques like RLHF~\cite{stiennon2020learning, ouyang2022training} and RLAIF~\cite{bai2022constitutional, lee2023rlaif} typically train heuristic reward models on pairwise rankings using the Bradley-Terry model~\cite{bradley1952rank}. While successful for general alignment, these scalar models lack formal guarantees w.r.t.\ preserving task-critical information and struggle with long-horizon credit assignment. While MaxIM is similar to RLAIF in its use of an oracle AI for offline data generation, we depart from preference-based rewards. Instead, we frame context compression as a reward shaping problem~\cite{ng1999policy}. Specifically, by training a critic to compute calibrated VPL bounds, MaxIM provides dense, step-by-step information-theoretic rewards. This directly optimizes rigorous MI lower bounds, promoting predictive sufficiency and factual grounding without the theoretical liabilities of standard preference approximations.

\paragraph{RL with Asymmetric Context.}
RL with asymmetric context (or privileged information) trains policies using a critic that has access to oracle information unavailable to the actor~\cite{pinto2017asymmetric, chen20a-learning-by-cheating}. Widely used in robotics to train student policies with expert teachers, we adapt it to LLM memory. By treating future text or latent intents as ``oracle hindsight context,'' we compute dense information-theoretic rewards that shape the LLM's summary update policy without expensive human labels.

\section{Background}
\label{sec:background}

\subsection{Capacity-Constrained Memory and Sparse Rewards}
\paragraph{Incremental Summarization.}
We formalize the environment as a sequential interaction stream. An
agent (or \emph{actor}) engages with an environment, generating actions $a_t \in \mathcal{A}$ given a continuous stream of observations $o_t \in \Omega$. Optimal decision making generally requires conditioning on the current history $H_t = (o_1, a_1, \dots, a_{t-1}, o_t)$. Maintaining $H_t$ indefinitely is computationally intractable for LLMs due to their finite context windows, hence, the agent adopts a summarizer policy $\pi_\theta(z_t \mid z_{t-1}, a_{t-1}, o_t)$ that incrementally updates a textual summary $z_t \in \bigcup_{k=1}^K \mathcal{V}^k$, where $\mathcal{V}$ is the LLM vocabulary and $K$ is a maximum token budget. The agent's policy conditions its action choice solely on the capacity-constrained working memory $z_t$ and the current observation $o_t$ to select its next action $a_t$.\footnote{We assume the agent conditions on memory and the latest observation, though it can use the last $k$ observations if compute and context permit.} We assume the agent is fixed/frozen and focus on training the summarizer for optimal performance.

\paragraph{Sparse Rewards and Regularized MDPs.}
The standard RL objective for optimizing policies maximizes expected cumulative task reward: $J(\theta) = \E_{\pi_\theta} [\sum_t \gamma^t R_{\text{task}, t}]$. However, in long-horizon settings, $R_{\text{task}}$ can be exceptionally sparse (e.g., evaluated only at the end of a multi-turn dialogue or after reading a full document). This sparsity makes RL challenging due to credit assignment failures. While classic reward shaping~\cite{ng1999policy} injects dense signals without altering the optimal policy via potential functions, this is insufficient in our capacity-constrained setting. Instead, we alter the objective to form a regularized Markov decision process (MDP)~\cite{geist2019theory}. By injecting dense, step-wise information-theoretic auxiliary rewards, we explicitly constrain the policy's exploration manifold, guiding it to find an effective bounded-memory policy rather than the theoretically infeasible unconstrained optimal (which would require unbounded memory).

\paragraph{A Unified POMDP Formulation.}
While our sequence compression formulation applies broadly, our framework models the interaction stream as a \emph{partially observable MDP (POMDP)}. This treats summary $z_t$ as a textual belief state tracking an unobservable latent state $s_t \in \mathcal{S}$. Crucially, this interpretation gracefully accommodates diverse streaming models, such as:\\
\textbf{Dynamic Latent States (e.g., User Simulation):} In multi-turn interactions, a user's underlying persona, intent, or mood acts as a dynamic latent state $s_t$ that evolves sequentially. The summary $z_t$ tracks this intent to guide the agent's conversational responses.\\
\textbf{Static Latent States and Active Perception (e.g., Goal-directed reading):} For tasks like reading a scientific paper to answer eventual questions, the environment state (i.e., full document semantics and golden target answers) is static. However, with bounded context, it becomes fundamentally partially observable. Intermediate actions (e.g., reading the next text chunk) are pure sensing actions, while the summary $z_t$ serves as the belief state, accumulating the necessary sufficient statistics from
the observations to resolve the unobserved target variables (questions) at the terminal step. By viewing both settings through this lens, we bridge incremental summarization and classical belief tracking, and thus establish formal value suboptimality bounds (see Section~\ref{sec:method}).

\subsection{Mutual Information and Representation Learning}
\paragraph{Estimating Mutual Information.}
Mutual information (MI), denoted $I(X; Y)$ and defined as $I(X; Y) = \E_{P(x,y)}\!\big[\log \frac{P(x,y)}{P(x)P(y)}\big]$, measures the reduction in uncertainty about a random variable $Y$ given knowledge of $X$. Estimating MI in high-dimensional continuous or discrete spaces is difficult. One prevalent approach in representation learning, \emph{contrastive predictive coding (CPC)}, uses the InfoNCE loss~\cite{oord2018representation}. While successfully applied to learning continuous embeddings, InfoNCE has been proven to be strictly upper-bounded by $\log N$, where $N$ is the batch size~\cite{mcallester2020formal}. This bound saturates rapidly when modeling high-entropy distributions like autoregressive text generation, limiting its utility as a metric for unbounded optimization.

\paragraph{Variational Bounds.}
To overcome this $\log N$ limitation, variational approximations based on $f$-divergences are often adopted, such as the NWJ~\cite{nguyen2007estimating} or SMILE~\cite{song2019understanding} estimators. However, these estimators are known to suffer from high variance, numerical instability, and extreme hyperparameter sensitivity, particularly when the true MI is large~\cite{poole2019variational}. Because of this, we focus on the VPL bound~\cite{barber2003vpl}. By introducing a parameterized variational distribution $q_\phi(Y \mid X)$, we leverage the non-negativity of KL divergence to establish a theoretically sound lower bound:
\begin{equation}
\label{eq:vpl-bound}
I(X; Y) \ge H(Y) + \E_{P(x,y)}[\log q_\phi(y \mid x)].
\end{equation}
In standard representation learning, maximizing this log-likelihood maximizes $I(X; Y)$ when the marginal entropy $H(Y)$ is constant. In closed-loop POMDPs, the agent's actions influence future states, causing $H(Y_t)$ to become weakly policy-dependent. However, as we prove in Section~\ref{sec:method}, our framework circumvents this dependency: maximizing predictive log-likelihood directly bounds conditional entropy $H(Y_t \mid z_t)$, which bounds the sequential information loss regardless of the marginal $H(Y_t)$. This provides a tractable, theoretically rigorous, low-variance surrogate objective that scales naturally to modern LLMs.

\section{Methodology: The MaxIM Framework}
\label{sec:method}

We frame capacity-constrained incremental summarization as a task of information-theoretic reward shaping. Our objective is to learn a \emph{summarizer update policy} $\pi_\theta$ that dynamically creates and maintains a working memory $z_t$, bounded by a strict token limit $K$, to maximize downstream task utility $R_{\text{task}}$. Crucially, $\pi_\theta$ is solely responsible for memory management; a separate, frozen action agent (e.g., a conversational assistant or a question-answering module) conditions on this memory $z_t$ to act within the environment. Because $R_{\text{task}}$ is often sparse and delayed in long-horizon streams (e.g., assessed only at the terminal turn $T$), standard RL struggles with credit assignment. Our goal is to derive principled, dense shaping rewards that offer step-by-step auxiliary guidance through the optimization landscape.

\subsection{Value Suboptimality and Predictive Sufficiency}
\label{subsec:sufficiency}
    Maintaining the full observation history $H_t = (H_{t-1}, a_{t-1}, o_t)$ is computationally intractable in streaming environments, forcing systems to compress $H_t$ into a bounded summary $z_t$. To quantify the impact of this compression on downstream task utility, we frame the sequential interaction as a POMDP, treating $z_t$ as a \emph{capacity-constrained Markovian state abstraction}~\cite{li2006towards, subramanian2022ais}. Let $Y_t$ denote the task-relevant target variable at step $t$---e.g., static comprehension answers in goal-directed reading ($Y_{t+1} = Y_t$), or a user's evolving latent state $s_t$ in multi-turn dialogue---and let $\mu_\theta$ denote the trajectory distribution induced by
    $\pi_\theta$ with the frozen action agent $\pi_{\text{act}}(a_t \mid z_t, o_t)$. As detailed in Appendix~\ref{app:proof_thm1}, we assume: \textbf{(A1) Target Mediation}, where expected task rewards and next-step state-observation transitions depend on $(H_t, a_t)$ through $(Y_t, a_t)$ (covering both static targets and closed-loop settings where $a_t$ alters $Y_{t+1}$); \textbf{(A2) Memory Causality} ($z_t \perp Y_t \mid H_t$), meaning $\pi_\theta$ updates $z_t$ from observed history without privileged access to $Y_t$; and \textbf{(A3) Compressed Action Selection} ($a_t \perp Y_t \mid H_t, z_t$), meaning the actor selects $a_t$ based on $(z_t, o_t)$ alone.

\begin{theorem}[Value Suboptimality Bound; extending~\cite{li2006towards, subramanian2022ais}]
\label{thm:suboptimality}
Let $V^*(H_t)$ be the optimal expected return achievable with the full observation history, and $V^*(z_t)$ be the optimal expected return achievable using the capacity-constrained summary $z_t$. Under Assumptions \textbf{(A1)}--\textbf{(A3)} with rewards bounded in $[-R_{\max}, R_{\max}]$, the expected suboptimality gap under the induced trajectory distribution $\mu_\theta$ is bounded by the sequential information loss $I_{\mu_\theta}(Y_t; H_t \mid z_t)$:
\begin{equation}
\label{eq:suboptimality}
\E_{\mu_\theta}\big[|V^*(H_t) - V^*(z_t)|\big] \le \frac{\sqrt{2}R_{\max}}{(1-\gamma)^2}\sqrt{\sup_k I_{\mu_\theta}(Y_k; H_k \mid z_k)}.
\end{equation}
\end{theorem}

Theorem~\ref{thm:suboptimality} (proved alongside an approximate-mediation extension in Appendix~\ref{app:proof_thm1}) establishes that to minimize the value suboptimality gap, it suffices for the agent to minimize sequential information loss $I_{\mu_\theta}(Y_t; H_t \mid z_t)$. This observation directly yields our first shaping objective:

\begin{proposition}[Predictive Sufficiency]
\label{prop:sufficiency}
To minimize the value suboptimality gap under a token budget $|z_t| \le K$, the policy can instead maximize the negative conditional entropy $-H_{\mu_\theta}(Y_t \mid z_t)$, which we call \emph{predictive sufficiency}. Introducing a parameterized variational distribution (critic) $q_\phi(Y_t \mid z_t)$, under \textbf{(A2)} sequential information loss satisfies $I_{\mu_\theta}(Y_t; H_t \mid z_t) = \E_{\mu_\theta}[-\log q_\phi(Y_t \mid z_t)] - H_{\mu_\theta}(Y_t \mid H_t) - \Delta_{\mathrm{post}}(\phi; \theta)$, where $\Delta_{\mathrm{post}}(\phi; \theta) \coloneqq \E_{z_t \sim \mu_\theta}[\KL(P_{\mu_\theta}(Y_t \mid z_t) \,\|\, q_\phi(Y_t \mid z_t))] \ge 0$. Thus, maximizing $\E_{\mu_\theta}[\log q_\phi(Y_t \mid z_t)]$ maximizes a variational lower bound on predictive sufficiency---minimizing an upper bound on $I_{\mu_\theta}(Y_t; H_t \mid z_t)$ up to closed-loop filtering-entropy shifts (zero for static targets; Appendix~\ref{app:proof_prop1})---without requiring constant marginal entropy $H_{\mu_\theta}(Y_t)$.
\end{proposition}

We prove Proposition~\ref{prop:sufficiency} in Appendix~\ref{app:proof_prop1}, where Proposition~\ref{prop:capacity_transfer} characterizes how budget $K$ and critic quality enter Eq.~\eqref{eq:suboptimality}: since $|z_t| \le K$ carries at most $K\log|\mathcal{V}| + \log 2$ nats, information loss obeys a rate-distortion floor decreasing with $K$ until $z_t$ is sufficient for $Y_t$. Maximizing $\E_{\mu_\theta}[\log q_\phi(Y_t \mid z_t)]$ approaches this frontier up to $\Delta_{\mathrm{post}}(\phi; \theta)$.

\subsection{Factual Grounding and Recursive Consistency}
\label{subsec:consistency}
While maximizing predictive sufficiency $-H_{\mu_\theta}(Y_t \mid z_t)$ ensures the summary retains task-relevant information, optimizing this objective in isolation can introduce a severe failure mode for LLMs:

\begin{lemma}[The Hallucination Failure Mode]
\label{lem:hallucination}
Let $\pi_\theta$ be a stochastic policy operating under token budget $K$. Because the mapping from a complex history $H_t$ to a sparse downstream target $Y_t$ is many-to-one, optimizing solely for predictive sufficiency $\max_\theta [-H_{\mu_\theta}(Y_t \mid z_t)]$ under-specifies the summary mapping. The policy can achieve the objective by fabricating plausible but ungrounded text that correlates with $Y_t$, sampled from its pre-trained prior. This maximizes the ungrounded conditional entropy $H(z_t \mid H_t)$ and severs the summary's dependence on the true history.
\end{lemma}

As argued in Appendix~\ref{app:proof_lemma1_halluc}, this lemma implies that an unconstrained policy can satisfy the predictive sufficiency objective by generating idealized, highly probable text that encodes the target $Y_t$ while suppressing $I(z_t; H_t)$ toward the minimum required by the data processing inequality (DPI). This vulnerability aligns with uncertainty quantification research in LLMs~\cite{kuhn2023semantic, manakul2023selfcheckgpt} showing that high ungrounded conditional entropy $H(z_t \mid H_t)$ is a primary indicator of confabulation.

\begin{proposition}[Recursive Consistency (Informal)]
\label{prop:consistency}
To suppress hallucination and promote factual grounding during streaming updates from local context $\tilde{H}_t = (z_{t-1}, a_{t-1}, o_t)$, the policy can regularize updates by maximizing the MI between the new summary and its conditioning context: $\max_\theta I_{\mu_\theta}(z_t; \tilde{H}_t)$. We refer to this regularizer as \emph{recursive consistency}.
\end{proposition}

We formalize and prove Proposition~\ref{prop:consistency} in Appendix~\ref{app:proof_prop2}. Since $I_{\mu_\theta}(z_t; \tilde{H}_t) = H_{\mu_\theta}(z_t) - H_{\mu_\theta}(z_t \mid \tilde{H}_t)$, maximizing $I_{\mu_\theta}(z_t; \tilde{H}_t)$ penalizes ungrounded entropy $H_{\mu_\theta}(z_t \mid \tilde{H}_t)$, while predictive sufficiency prevents $z_t$ from collapsing into a trivial constant summary (which would drop $-H_{\mu_\theta}(Y_t \mid z_t)$ back to $-H_{\mu_\theta}(Y_t)$). Because $\pi_\theta$ observes only $\tilde{H}_t$, any task-relevant detail omitted from $z_{t-1}$ is permanently lost unless re-observed: Proposition~\ref{prop:markov_tracking} and Corollary~\ref{cor:myopic_dpi} (Appendix~\ref{app:proof_prop2}) formalize how per-turn losses accumulate, vanishing when $K$ suffices to update the task-relevant sufficient statistics from $\tilde{H}_t$ and showing why myopic single-step compression fails.

\subsection{The MaxIM Objective and Unified Asymmetric Critic Training}
\label{subsec:objective}
We operationalize Propositions~\ref{prop:sufficiency} and~\ref{prop:consistency} into step-wise RL rewards using the local streaming context $\tilde{H}_t = (z_{t-1}, a_{t-1}, o_t)$. Computing exact MI online during RL rollouts is intractable; instead, we generate an offline synthetic dataset $\mathcal{D}$ with privileged access to $Y_t$ (Appendix~\ref{app:data_generation}) and pre-train a critic $q_\phi$ via teacher forcing.

\paragraph{From Sufficiency and Consistency to Pointwise Mutual Information (PMI) Rewards.}
To turn both objectives into calibrated step-wise rewards, we cast each as a log-likelihood ratio. For \emph{predictive sufficiency} (Proposition~\ref{prop:sufficiency}), using $\log q_\phi(Y_t \mid z_t)$ as-is conflates summary quality with target difficulty: easy targets score well even when $z_t$ is uninformative, whereas hard targets score poorly even for thorough summaries. Subtracting a summary-free baseline $\log q_\phi(Y_t)$---independent of $z_t$ conditional on $(H_t, Y_t)$, leaving the immediate policy gradient of $\log q_\phi(Y_t \mid z_t)$ unbiased---isolates the information $z_t$ provides about $Y_t$ in PMI form, $\log \frac{q_\phi(Y_t \mid z_t)}{q_\phi(Y_t)}$. For \emph{recursive consistency} (Proposition~\ref{prop:consistency}), $I(z_t; \tilde{H}_t) = \E\!\big[\log \frac{P(\tilde{H}_t \mid z_t)}{P(\tilde{H}_t)}\big] = \E\!\big[\log \frac{P(z_t \mid \tilde{H}_t)}{P(z_t)}\big]$. Scoring consistency via \emph{backward} reconstruction $\log q_\phi(\tilde{H}_t \mid z_t)$ fails to penalize hallucinated claims appended alongside true statements, since fabricated sentences in $z_t$ need not lower the likelihood of reconstructing $\tilde{H}_t$. Estimating $I(z_t; \tilde{H}_t)$ in the \emph{forward} direction, $\log \frac{P(z_t \mid \tilde{H}_t)}{P(z_t)}$, evaluates each token of $z_t$: claims unsupported by $\tilde{H}_t$ are no more probable given $\tilde{H}_t$ than unconditionally, yielding zero or negative log-ratios.

Because both objectives share this conditional-to-unconditional log-ratio structure, at each step $t$ we query $q_\phi$ with the conditioning input present and replaced by a fixed null placeholder---$z_{\mathrm{null}}$ when omitting $z_t$ (e.g., \texttt{"(no summary available)"}) to approximate $P(Y_t)$, and $\tilde{H}_{\mathrm{null}}$ when omitting $(z_{t-1}, a_{t-1}, o_t)$ to approximate $P(z_t)$. Routing all four queries via mode-specific system instructions (Appendix~\ref{app:critic_prompts}) trains a \textbf{single unified critic} $q_\phi$ via cross-entropy loss on $\mathcal{D}$:
\begin{equation}
\label{eq:critic_loss}
\begin{aligned}
\mathcal{L}(\phi) = -\E_{\mathcal{D}}\Big[&\lambda_{\text{pred}}\big(\log q_\phi(Y_t \mid z_t) + \log q_\phi(Y_t \mid z_{\mathrm{null}})\big) \\[-2pt]
&+ \lambda_{\text{cons}}\big(\log q_\phi(z_t \mid \tilde{H}_t) + \log q_\phi(z_t \mid \tilde{H}_{\mathrm{null}})\big)\Big],
\end{aligned}
\end{equation}
with loss weights $\lambda_{\text{pred}}, \lambda_{\text{cons}} > 0$. During RL, the frozen $q_\phi$ scores $z_t$ at turn $t$ via length-normalized \textbf{PMI} using teacher-forced forward passes only (zero critic generation):
\begin{align}
\label{eq:r_pred}
r_{\text{pred}, t} &= \frac{1}{|Y_t|} \sum_{i=1}^{|Y_t|} \Big[ \log q_\phi(Y_{t,i} \mid Y_{t,<i}, z_t) - \log q_\phi(Y_{t,i} \mid Y_{t,<i}, z_{\mathrm{null}}) \Big], \\
\label{eq:r_cons}
r_{\text{cons}, t} &= \frac{1}{|z_t|} \sum_{i=1}^{|z_t|} \Big[ \log q_\phi(z_{t,i} \mid z_{t,<i}, \tilde{H}_t) - \log q_\phi(z_{t,i} \mid z_{t,<i}, \tilde{H}_{\mathrm{null}}) \Big].
\end{align}
Proposition~\ref{prop:pmi_nonstationarity} (Appendix~\ref{app:proof_prop3}) formalizes per-step baseline unbiasedness, multi-step/sigmoid trade-offs, the KL decomposition for $r_{\text{cons}, t}$, and critic shift bounds (validated in Appendix~\ref{app:critic_probes}).

\subsection{Policy Update and Algorithm Overview}
\label{subsec:policy_update}
During RL, the summarizer policy maps $(z_{t-1}, a_{t-1}, o_t)$ to an updated summary $z_t$. Because discarding critical information early causes irreversible downstream loss, we propagate future task and shaping signals back to each turn $t$ using a backward-discounted return-to-go $G_{t,s}$ at RL step $s$.

We map $r_{\text{pred}, t}$ and $r_{\text{cons}, t}$ onto a common bounded scale via centered sigmoids:
\begin{equation}
\label{eq:shaped_step}
\hat{r}_t = w_s \big(\sigma(\alpha_s(r_{\text{pred}, t} - b_s)) - \tfrac{1}{2}\big) + w_c \big(\sigma(\alpha_c(r_{\text{cons}, t} - b_c)) - \tfrac{1}{2}\big).
\end{equation}
Here $(w_s, w_c)$ weighs sufficiency and consistency, respectively, with $w_c < w_s$ so consistency acts as a grounding regularizer and does not dominate sufficiency; and $(\alpha_s, b_s, \alpha_c, b_c)$ are calibrated offline using summaries sampled from the initial pre-RL base policy (Appendix~\ref{app:implementation-details}). Centering at $-\tfrac{1}{2}$ assigns baseline-quality summaries zero shaping reward (avoiding positive length bias on variable-horizon episodes). Because $\hat{r}_k$ is summed across turns $k = t, \dots, T$ whereas $R_{\text{task}}$ is sparse, we scale $\hat{r}_k$ by $\eta_{\text{shape}} = 1 - \gamma$ so accumulated shaping rewards remain balanced against $R_{\text{task}}$. Finally, a cosine schedule $\lambda_s \in [0, 1]$ over RL update iterations $s$ holds $\lambda_s = 1$ early to resolve credit assignment and decays $\lambda_s \to 0$ mid-training so late-stage updates optimize $R_{\text{task}}$ without residual critic bias:
\begin{equation}
\label{eq:return_to_go}
G_{t,s} = \sum_{k=t}^T \gamma^{k-t} \big( R_{\text{task}, k} + \eta_{\text{shape}} \lambda_s\, \hat{r}_k \big).
\end{equation}
We update $\pi_\theta$ using KL-regularized PPO~\cite{schulman2017proximal} with group-based relative advantage estimation~\cite{shao2024deepseekmath}: for each initial state, $G$ rollouts are sampled and $G_{t,s}$ is standardized at each turn $t$ via a leave-one-out baseline across the remaining $G-1$ active rollouts in the group (Algorithm~\ref{alg:maxim}, Appendix~\ref{app:algorithm}).

\section{Experiments}
\label{sec:experiments}

We evaluate MaxIM on two complementary streaming benchmarks instantiating our static and dynamic POMDP formulations (Section~\ref{sec:method}), addressing three research questions:
\begin{itemize}[leftmargin=1.4em, itemsep=1pt, topsep=2pt]
    \item \textbf{RQ1:} Does MaxIM's information-theoretic reward shaping improve downstream task utility and token-budget Pareto efficiency over state-of-the-art memory baselines?
    \item \textbf{RQ2:} How does MaxIM perform on fine-grained atomic-claim grounding and qualitative axes---Truthfulness, Usefulness, and Consistency---under cross-family LLM judges?
    \item \textbf{RQ3:} How do MaxIM's individual reward components (sufficiency PMI, recursive consistency PMI, and mid-training cosine annealing) contribute to task utility and grounding?
\end{itemize}

\subsection{Experimental Setup}
\label{subsec:setup}

\paragraph{Tasks and Environments.}
\textbf{Task 1: arXiv Scientific Reading.} The agent incrementally reads 500-token chunks of a paper (up to $20$ in RL; $25$ at test) under a strict $K=500$ token budget, plus $K \in \{100, 300, 800\}$ to test budget scaling. To minimize pretraining leakage, we curate $2{,}196$ recent ar5iv papers~\cite{ar5iv} across nine domains into disjoint critic/SFT ($1{,}281$), RL ($533$), validation ($44$), and post-cutoff test ($338$) splits, each with five cross-section synthesis multiple choice questions (MCQs) (Appendix~\ref{app:arxiv_data}). To penalize guessing with no evidence and to evaluate evidence explicitly retained in $z_T$, we use \textbf{Abstain-MCQ Accuracy}: given only $z_T$, the frozen actor selects A--D or E (\emph{``Insufficient information''}), earning $+1$ for correct, $0$ for E, and $-1/3$ for wrong guesses ($0$ random expectation). An \textbf{Early-Section Stress Test} ($332$ held-out papers) probes retention of technical details (design justifications, ablation trade-offs) appearing only in chunks $3$--$7$ across $13$--$22$ subsequent updates.

\textbf{Task 2: WildChat Multi-Turn User Simulation.} The agent tracks a user's evolving latent state under a $K=1{,}024$ token budget using a 10-key JSON belief schema (goals, preferences, constraints, affect) in multi-turn conversations. From English WildChat dialogues~\cite{zhao2024wildchat} ($\ge 6$ user turns), we annotate causal per-turn states offline, train a user simulator to generate utterances and ground-truth states $s_t$ (Appendix~\ref{subsec:user_sim}), and form disjoint critic/SFT ($3{,}000$), RL ($3{,}500$), validation ($200$), and test ($500$) splits (Appendix~\ref{app:wildchat_data}). Because conversational quality spans multiple axes that a single online scalar can misspecify, RL optimizes terminal belief-tracking accuracy ($1$--$10$ rating rescaled to $R_{\text{task}} \in [0, 1]$; Appendix~\ref{app:wc_belief_fidelity}), whereas test evaluation measures multi-axis agent utility in fresh closed-loop rollouts ($T \le 15$). 
Specifically on the test dialogues, the frozen assistant sees only a local window ($k=1$) with just the most recent turn $(u_{t-1}, a_{t-1})$, the latest user message $u_t$, and $z_{t-1}$. 
Given $(u_{1:T}, a_{1:T})$ and true states $s_{0:T}$, GPT-4.1 rates ($1$--$10$) \textbf{Overall User Satisfaction}, \textbf{Goal Completion}, \textbf{Preference \& Constraint Retention}, and \textbf{Coherence \& Adaptability} (Appendix~\ref{app:compute_cost}). Table~\ref{tab:main_rq1} reports signed $\Delta$ vs.\ a memory-free \textbf{No-Belief} baseline ($k=1$ only) on all $N=500$ dialogues and on $T \ge 10$ dialogues ($N=158$ seeds with $\ge 10$ turns under No-Belief).

\paragraph{Baselines and Implementation.}
We compare MaxIM against three non-RL baselines: \textbf{Prompt-Only} (zero-shot base updates), \textbf{Policy SFT} (distilled on offline oracle summaries; Appendix~\ref{app:data_generation}), and \textbf{Best-of-$N$ MI SFT} ($N=8$ on arXiv; distilling single-step updates with highest shaping reward $\hat{r}_t$ to test myopic MI selection); two sparse-reward RL baselines: \textbf{MemAgent RL}~\cite{yu2025memagent} and \textbf{Task-Only RL}, which do not use MI shaping; and two ablations: \textbf{Suff-Only RL} ($w_c=0$, no recursive consistency) and \textbf{MI-Only RL} ($R_{\text{task}}=0$, shaping only). Summarizers, the unified critics, frozen actors, and the user simulator use \textbf{Gemini 2.5 Flash}~\cite{comanici2025gemini25pushingfrontier}; offline summary target synthesis and the RL belief rater use \textbf{Gemini 3.0 Flash}; and test-time WildChat ratings, atomic-claim grounding, and qualitative evaluations use cross-family \textbf{GPT-4.1}. RL methods use $G=8$ and $\gamma=0.9$ ($\gamma=1.0$ for MemAgent RL), with zero critic calls at test time.

\begin{table}[t]
\centering
\caption{\textbf{Main downstream task utility results (RQ1).} \emph{Left:} arXiv Abstain-MCQ accuracy ($\pm$ SEM; $N=338$), abstention rate, and Early-Section Stress Test ($N=332$, chunks $3$--$7$). \emph{Right:} WildChat paired $\Delta$ vs.\ memory-free \textbf{No-Belief} ($k=1$ only) on $1$--$10$ GPT-4.1 agent ratings over all $N=500$ dialogues (No-Belief: $8.58$ Sat., $8.75$ Goal, $8.74$ Pref.) and the $T \ge 10$ cohort ($N=158$; No-Belief: $8.08$ Sat., $8.28$ Pref.; full scores and Coherence in Table~\ref{tab:wc_downstream_full}). Best \textbf{bold}; second-best \underline{underlined}. $^{\dagger}p < 0.05$, $^{\ddagger}p < 0.01$ vs.\ MaxIM (paired bootstrap).}
\label{tab:main_rq1}
\vspace{2pt}
{\small
\setlength{\tabcolsep}{3.6pt}
\begin{tabular}{lcccccccc}
\toprule
& \multicolumn{3}{c}{\textbf{arXiv Scientific Reading ($K{=}500$)}} & \multicolumn{5}{c}{\textbf{WildChat Dialogue ($\Delta$ vs.\ No-Belief $\uparrow$)}} \\
\cmidrule(lr){2-4} \cmidrule(lr){5-9}
\multirow{2}{*}[-2pt]{\textbf{Method}} & \multirow{2}{*}[-2pt]{\shortstack{\textbf{Abstain}\\\textbf{MCQ $\uparrow$}}} & \multirow{2}{*}[-2pt]{\shortstack{\textbf{Abst.}\\\textbf{(\%) $\downarrow$}}} & \multirow{2}{*}[-2pt]{\shortstack{\textbf{Early}\\\textbf{Chunk $\uparrow$}}} & \multicolumn{3}{c}{\textbf{All $T$ ($N{=}500$)}} & \multicolumn{2}{c}{\textbf{$T{\ge}10$ ($N{=}158$)}} \\
\cmidrule(lr){5-7} \cmidrule(lr){8-9}
& & & & \textbf{Sat.} & \textbf{Goal} & \textbf{Pref.} & \textbf{Sat.} & \textbf{Pref.} \\
\midrule
Prompt-Only & $31.30{\scriptstyle\pm 1.28}$ & $68.1$ & $17.35{\scriptstyle\pm 1.05}$ & $-0.09$ & $-0.09$ & $+0.01$ & $-0.09$ & $+0.08$ \\
Policy SFT & $38.76{\scriptstyle\pm 1.29}$ & $59.1$ & $12.47{\scriptstyle\pm 0.87}$ & $-0.04$ & $-0.05$ & $+0.05$ & $-0.21$ & $\phantom{+}0.00$ \\
Best-of-$N$ MI SFT & $29.05{\scriptstyle\pm 1.16}$ & $70.1$ & --- & --- & --- & --- & --- & --- \\
MemAgent RL~\cite{yu2025memagent} & $52.90{\scriptstyle\pm 1.40}$ & $46.0$ & --- & $-0.28^{\ddagger}$ & $-0.27^{\ddagger}$ & $-0.13^{\ddagger}$ & $-0.15$ & $+0.02$ \\
Task-Only RL & \underline{$54.44{\scriptstyle\pm 1.45}$} & \underline{$43.7$} & \underline{$32.23{\scriptstyle\pm 1.26}$} & $-0.30^{\ddagger}$ & $-0.29^{\ddagger}$ & $-0.15^{\ddagger}$ & $-0.37^{\dagger}$ & $-0.11^{\dagger}$ \\
\midrule
MI-Only RL & $50.30{\scriptstyle\pm 1.50}$ & $47.8$ & --- & $\phantom{+}\underline{0.00}$ & $\phantom{+}\underline{0.00}$ & \underline{$+0.07$} & $\mathbf{+0.18}$ & \underline{$+0.23$} \\
Suff-Only RL & $51.54{\scriptstyle\pm 1.46}$ & $46.9$ & --- & $-0.18^{\dagger}$ & $-0.14$ & $-0.12^{\ddagger}$ & \underline{$+0.16$} & \underline{$+0.23$} \\
\textbf{MaxIM (Ours)} & $\mathbf{66.39{\scriptstyle\pm 1.31}}$ & $\mathbf{32.0}$ & $\mathbf{48.55{\scriptstyle\pm 1.35}}$ & $\mathbf{+0.05}$ & $\mathbf{+0.05}$ & $\mathbf{+0.16}$ & $+0.11$ & $\mathbf{+0.30}$ \\
\bottomrule
\end{tabular}%
}
\vspace{-6pt}
\end{table}

\subsection{Main Results: Downstream Task Utility and Budget Scaling (RQ1)}
\label{subsec:rq1}

\paragraph{Scientific Reading and Long-Range Credit Assignment.}
On arXiv ($K=500$, Table~\ref{tab:main_rq1}), non-RL baselines abstain on $59\%$--$70\%$ of synthesis questions: Prompt-Only ($31.30\%$) lacks guidance on which details matter downstream, Policy SFT ($38.76\%$) compounds exposure bias over $25$ updates, and \textbf{Best-of-$N$ MI SFT} ($29.05\%$) confirms Corollary~\ref{cor:myopic_dpi} (Appendix~\ref{app:proof_prop2})---greedily selecting single-step updates by immediate PMI cannot plan capacity across chunks or replace non-myopic RL. Sparse-reward RL substantially improves retention (\textbf{MemAgent} $52.90\%$, \textbf{Task-Only} $54.44\%$), yet \textbf{MaxIM} reaches $\mathbf{66.39\%}$, beating Task-Only by $\mathbf{+11.95}$~pp and MemAgent by $\mathbf{+13.49}$~pp and cutting abstention to $32.0\%$. On the \textbf{Early-Section Stress Test} (chunks $3$--$7$), MaxIM's lead widens to $\mathbf{48.55\%}$ vs.\ $32.23\%$ for Task-Only ($\mathbf{+50.6\%}$ relative) and $17.35\%$ for Prompt-Only, confirming dense PMI shaping helps retain early task-relevant details.

\begin{figure}[t]
    \centering
    \includegraphics[width=\textwidth]{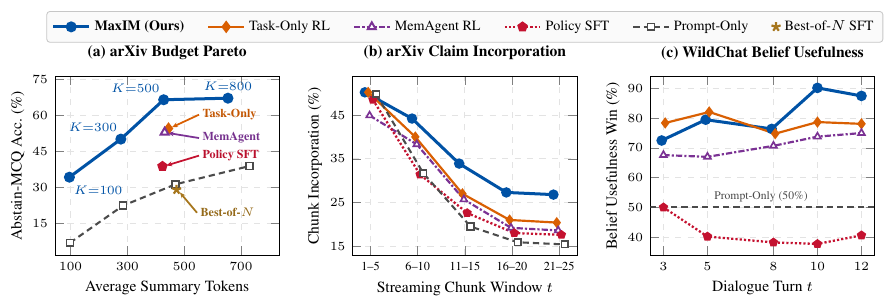}
    \vspace{-12pt}
    \caption{\textbf{Budget scaling and long-horizon dynamics.} \textbf{(a)} arXiv token-budget Pareto curve ($K \in \{100,300,500,800\}$; $K=500$ baselines labeled): MaxIM at $K=100$ outperforms Prompt-Only at a $5\times$ larger $K=500$ budget. \textbf{(b)} Chunk claim incorporation (percentage of atomic claims in chunk $o_t$ entailed by $z_t$) across 5-chunk windows ($t=1\dots 25$): MaxIM resists late-turn memory saturation. \textbf{(c)} WildChat GPT-4.1 pairwise belief-state Usefulness win rate vs.\ Prompt-Only.}
    \label{fig:scaling}
    \vspace{-8pt}
\end{figure}

\paragraph{Token-Budget Pareto Frontier.}
Across strict budgets $K \in \{100, 300, 500, 800\}$ (Figure~\ref{fig:scaling}(a)), MaxIM dominates the Pareto frontier: at $K=100$, MaxIM achieves $\mathbf{34.26\%}$ Abstain-MCQ---beating Prompt-Only at a $5\times$ larger $K=500$ budget ($31.30\%$)---and at $K=500$ ($66.39\%$) it surpasses Prompt-Only even at $K=2{,}000$ ($64.32\%$; Table~\ref{tab:e2_full}), saturating smoothly at $67.04\%$ ($K=800$) per Proposition~\ref{prop:capacity_transfer}. Figure~\ref{fig:scaling}(b) tracks \emph{chunk claim incorporation} (the share of atomic claims in chunk $o_t$ entailed by $z_t$): all methods incorporate $45\%$--$50\%$ in chunks $1$--$5$, but baselines saturate once the $K=500$ buffer fills whereas MaxIM sustains $\mathbf{26.8\%}$ even in chunks $21$--$25$.

\paragraph{Bounded-Memory Multi-Turn Dialogue: When Imperfect Beliefs Hurt.}
On WildChat (Table~\ref{tab:main_rq1}), the assistant with no belief tracking (No-Belief) already scores high on average ($8.58$ Sat., $8.74$ Pref.). However, its scores drop noticeably on longer dialogues ($T \ge 10$) to $8.08$ and $8.28$. Strikingly, \textbf{every baseline without MI shaping underperforms No-Belief} on Satisfaction and Goal Completion---including \textbf{Task-Only RL} ($\Delta = -0.30$, $p=0.0084$; $-0.37$ on $T \ge 10$) and \textbf{MemAgent RL} ($\Delta = -0.28$, $p=0.0206$): conditioning replies on inaccurately tracked beliefs steers the assistant toward false assumptions, proving worse than no belief state at all. Notably, Task-Only ($8.79/10$) and MemAgent ($8.98/10$) nearly match MaxIM ($9.00/10$; Table~\ref{tab:wc_belief_fidelity}) on the terminal belief-tracking reward $R_{\text{task}}$ itself; optimizing terminal belief accuracy alone fails to improve agent utility because it tolerates intermediate-turn hallucinations and omissions that derail live responses (Section~\ref{subsec:autorater}).

\textbf{Long-Horizon Recovery via MI Shaping.} By contrast, on $T \ge 10$ dialogues, \textbf{only MI-shaped policies outperform No-Belief} across every metric (bottom block of Table~\ref{tab:main_rq1}). Full \textbf{MaxIM} is the \emph{only} method with positive gains ($\Delta > 0$) across all metrics on both cohorts, hitting $\Delta = \mathbf{+0.30}$ on $T \ge 10$ Preference Retention while resolving goals in fewer turns ($10.68$ vs.\ $12.61$; Table~\ref{tab:wc_downstream_full}), beating Task-Only by $\mathbf{+0.35}$ Sat.\ ($p=0.0012$; $\mathbf{+0.48}$ on $T \ge 10$) and MemAgent by $\mathbf{+0.34}$ ($p=0.0032$).

\subsection{Factual Grounding and Qualitative Assessment (RQ2)}
\label{subsec:autorater}

\begin{table}[t]
\centering
\caption{\textbf{Source grounding and qualitative summary evaluation (RQ2; GPT-4.1).} Grounding of terminal $z_T$, pairwise \textbf{Usefulness}/\textbf{Truthfulness} win rates vs.\ Prompt-Only, and $1$--$10$ \textbf{Consistency}.}
\label{tab:grounding_rq2}
\vspace{2pt}
{\footnotesize
\setlength{\tabcolsep}{4.2pt}
\begin{tabular}{lcccccccccc}
\toprule
& \multicolumn{5}{c}{\textbf{arXiv Scientific Reading ($N{=}338$)}} & \multicolumn{5}{c}{\textbf{WildChat Multi-Turn Dialogue ($N{=}500$)}} \\
\cmidrule(lr){2-6} \cmidrule(lr){7-11}
\textbf{Method} & \shortstack{\textbf{Entail}\\\textbf{(\%) $\uparrow$}} & \shortstack{\textbf{Contr}\\\textbf{(\%) $\downarrow$}} & \shortstack{\textbf{Gnd.}\\\textbf{Claims $\uparrow$}} & \shortstack{\textbf{Use}\\\textbf{Win \%}} & \shortstack{\textbf{Cons.}\\\textbf{(1--10)}} & \shortstack{\textbf{Entail}\\\textbf{(\%) $\uparrow$}} & \shortstack{\textbf{Contr}\\\textbf{(\%) $\downarrow$}} & \shortstack{\textbf{Gnd.}\\\textbf{Claims $\uparrow$}} & \shortstack{\textbf{Use}\\\textbf{Win \%}} & \shortstack{\textbf{Truth}\\\textbf{Win \%}} \\
\midrule
Prompt-Only & $98.04$ & $\mathbf{0.50}$ & $33.0$ & --- & $6.16$ & $84.38$ & $11.04$ & $18.7$ & --- & --- \\
Policy SFT & $75.01$ & $11.62$ & $21.0$ & $13.4$ & \underline{$7.12$} & $83.68$ & $9.00$ & $16.5$ & $37.6$ & $36.2$ \\
MemAgent RL & $98.35$ & $0.62$ & $47.6$ & $\mathbf{80.1}$ & $6.07$ & $85.42$ & $9.35$ & $21.0$ & $66.4$ & $61.0$ \\
Task-Only RL & \underline{$98.37$} & $0.65$ & $47.3$ & $62.5$ & $6.44$ & $87.50$ & $7.04$ & \underline{$23.1$} & $\mathbf{79.8}$ & \underline{$69.8$} \\
MI-Only RL & $98.13$ & $0.78$ & $\mathbf{63.4}$ & $47.0$ & $6.04$ & $85.19$ & $10.53$ & $20.5$ & $40.8$ & $40.8$ \\
Suff-Only RL & $\mathbf{98.39}$ & \underline{$0.56$} & $49.9$ & \underline{$74.4$} & $5.99$ & \underline{$89.77$} & \underline{$6.48$} & $22.0$ & $72.2$ & $65.8$ \\
\textbf{MaxIM (Full)} & $97.48$ & $1.06$ & \underline{$54.0$} & $72.9$ & $\mathbf{7.55}$ & $\mathbf{91.75}$ & $\mathbf{4.27}$ & $\mathbf{40.3}$ & \underline{$73.8$} & $\mathbf{72.4}$ \\
\bottomrule
\end{tabular}%
}
\vspace{-8pt}
\end{table}

GPT-4.1 evaluates two protocols (Table~\ref{tab:grounding_rq2}; Appendix~\ref{app:prompts}): \textbf{atomic-claim grounding} decomposes terminal $z_T$ into claims classified against the source as supported (\textbf{Entail \%}), neutral, or fabricated/distorted (\textbf{Contr \%}), and reports mean supported claims (\textbf{Gnd.\ Claims}); \textbf{qualitative evaluation} compares each method vs.\ Prompt-Only ($t=20$ on arXiv; $z_T$ on WildChat) on task helpfulness (\textbf{Use Win \%}) and faithfulness (\textbf{Truth Win \%}), and scores arXiv \textbf{Consistency} (\textbf{Cons.}) on a $1$--$10$ scale by whether $(z_{t-1}, o_t) \to z_t$ preserves prior context while integrating $o_t$.

\paragraph{Factual Grounding and Qualitative Trade-Offs on Scientific Reading.}
In Table~\ref{tab:grounding_rq2} (left), RL policies sustain high arXiv entailment ($97.5\%$--$98.4\%$), whereas Policy SFT degrades ($75.01\%$ entailment, $11.62\%$ contradiction) as errors compound across $25$ sequential updates. MaxIM attains $\mathbf{97.48\%}$ entailment while packing $\mathbf{54.0}$ \textbf{grounded claims}---$63.6\%$ above Prompt-Only ($33.0$) and above Task-Only ($47.3$) and MemAgent ($47.6$), whereas MI-Only packs $63.4$ claims with low selectivity ($47.0\%$ Usefulness). Against Prompt-Only at $t=20$, MemAgent has a higher Usefulness win rate than MaxIM ($80.1\%$ vs.\ $72.9\%$), whereas MaxIM achieves the highest Consistency overall ($\mathbf{7.55}$ vs.\ MemAgent's $6.07$), retains more claims ($54.0$ vs.\ $47.6$), and leads by $13.49$~pp on Abstain-MCQ. Moreover, when GPT-4.1 compares MaxIM and MemAgent directly ($t = 3\dots 20$), it prefers MaxIM on Usefulness at every chunk ($54.8\%$--$72.0\%$ win rate; Table~\ref{tab:turnwise_e4}).

\paragraph{Dynamic Belief Grounding on WildChat.}
While arXiv provides explicit text, WildChat requires inferring \emph{implicit} user states without confabulation. MaxIM achieves the highest entailment ($\mathbf{91.75\%}$), cuts contradictions to $\mathbf{4.27\%}$---$\mathbf{54\%}$ below MemAgent ($9.35\%$), $\mathbf{39\%}$ below Task-Only ($7.04\%$), and $\mathbf{61\%}$ below Prompt-Only ($11.04\%$)---and retains nearly \textbf{twice} as many grounded claims ($40.3$ vs.\ $21.0$--$23.1$), explaining why MaxIM aids the $k=1$ assistant while unshaped RL misleads it. MaxIM also leads Truthfulness ($\mathbf{72.4\%}$) and ranks second in Usefulness ($73.8\%$ vs.\ Task-Only's $79.8\%$). Figure~\ref{fig:scaling}(c) (Table~\ref{tab:turnwise_e4}) shows why: for $t \le 5$, Task-Only's speculative inferences edge MaxIM on Usefulness ($82.1\%$ vs.\ $79.5\%$ at $t=5$) at lower Truthfulness ($70.5\%$ vs.\ $\mathbf{78.6\%}$), whereas for $t \ge 8$ MaxIM leads every baseline on \emph{both} Usefulness (up to $\mathbf{90.2\%}$) and Truthfulness ($\mathbf{88.5\%}$).

\subsection{Ablation Studies: Reward Components and Annealing Dynamics (RQ3)}
\label{subsec:ablation}

\begin{table}[t]
\centering
\caption{\textbf{Reward-component and annealing ablations (RQ3).} \textbf{(a)} Isolating $R_{\text{task}}$, $r_{\text{pred}}$, and $r_{\text{cons}}$ (\textbf{WC $\Delta$ Sat.} vs.\ No-Belief). \textbf{(b)} Cosine-decay window of $\lambda_s$ (\% of RL steps; $N=338$).}
\label{tab:ablation_rq3}
\vspace{2pt}
{\small
\setlength{\tabcolsep}{2.0pt}
\begin{tabular}{ccc lccc @{\hspace{6pt}} lcc}
\toprule
\multicolumn{7}{c}{\textbf{(a) Reward Component Ablation}} & \multicolumn{3}{c}{\textbf{(b) arXiv Annealing ($K{=}500$)}} \\
\cmidrule(r){1-7} \cmidrule(l){8-10}
$R_{\text{task}}$ & $r_{\text{pred}}$ & $r_{\text{cons}}$ & \textbf{Variant} & \shortstack{\textbf{arXiv MCQ}\\\textbf{(\%) $\uparrow$}} & \shortstack{\textbf{WC Contr}\\\textbf{(\%) $\downarrow$}} & \shortstack{\textbf{WC}\\\textbf{$\Delta$ Sat. $\uparrow$}} & \shortstack{\textbf{Decay}\\\textbf{(\% Steps)}} & \shortstack{\textbf{Val MCQ}\\\textbf{(\%)}} & \shortstack{\textbf{Test MCQ}\\\textbf{(\%) $\uparrow$}} \\
\midrule
\checkmark & & & Task-Only & \underline{$54.44{\scriptstyle\pm 1.45}$} & $7.04{\scriptstyle\pm 0.49}$ & $-0.30$ & None ($\lambda_s{=}1$) & $73.18$ & $57.10{\scriptstyle\pm 1.40}$ \\
\checkmark & \checkmark & & Suff-Only & $51.54{\scriptstyle\pm 1.46}$ & \underline{$6.48{\scriptstyle\pm 0.41}$} & $-0.18$ & Early ($13$--$60\%$) & $73.18$ & \underline{$61.18{\scriptstyle\pm 1.33}$} \\
& \checkmark & \checkmark & MI-Only & $50.30{\scriptstyle\pm 1.50}$ & $10.53{\scriptstyle\pm 0.44}$ & $\phantom{+}\underline{0.00}$ & Late ($65$--$90\%$) & \underline{$74.55$} & $61.12{\scriptstyle\pm 1.44}$ \\
\checkmark & \checkmark & \checkmark & \textbf{MaxIM} & $\mathbf{66.39{\scriptstyle\pm 1.31}}$ & $\mathbf{4.27{\scriptstyle\pm 0.27}}$ & $\mathbf{+0.05}$ & \textbf{Mid ($45$--$60\%$)} & $\mathbf{75.91}$ & $\mathbf{66.39{\scriptstyle\pm 1.31}}$ \\
& & & \emph{vs.\ 2nd-best:} & $\mathbf{+11.95}$\,\textbf{pp} & $\mathbf{-2.21}$\,\textbf{pp} & $\mathbf{+0.05}$ & \emph{vs.\ none:} & $\mathbf{+2.73}$\,\textbf{pp} & $\mathbf{+9.29}$\,\textbf{pp} \\
\bottomrule
\end{tabular}%
}
\vspace{-8pt}
\end{table}

\paragraph{Synergy of Sufficiency, Recursive Consistency, and Task Anchoring.}
Table~\ref{tab:ablation_rq3}(a) isolates each reward term: only full MaxIM achieves positive WildChat $\Delta$ Sat.\ ($+0.05$; $+0.23$ vs.\ Suff-Only, $+0.35$ vs.\ Task-Only). Dropping recursive consistency (\textbf{Suff-Only RL}, $w_c=0$) cuts arXiv accuracy by $-14.85$~pp ($51.54\%$), yields the lowest consistency ($5.99$; Table~\ref{tab:grounding_rq2}), and raises WildChat contradictions by $+51.8\%$ ($6.48\%$ vs.\ $4.27\%$): while Suff-Only aids $T \ge 10$ dialogues ($\Delta = +0.16$; Table~\ref{tab:main_rq1}), unregularized sufficiency rewards speculative claims (Lemma~\ref{lem:hallucination}) that pull overall Satisfaction negative ($\Delta = -0.18$, $p=0.048$). Conversely, \textbf{MI-Only RL} ($R_{\text{task}}=0$) reaches $50.30\%$ arXiv accuracy and $+0.18$ $T \ge 10$ Satisfaction \emph{without task reward} (Proposition~\ref{prop:sufficiency}), yet leaves a $16.09$~pp arXiv gap and $10.53\%$ contradictions (Proposition~\ref{prop:pmi_nonstationarity}(ii)), motivating task anchoring.

\paragraph{Role of Mid-Training Cosine Annealing.}
Table~\ref{tab:ablation_rq3}(b) varies the $\lambda_s$ decay window on arXiv ($K=500$). Constant shaping (\textbf{No Annealing}, $57.10\%$) beats Task-Only ($54.44\%$), while decaying too early ($13\%$--$60\%$, $61.18\%$) or late ($65\%$--$90\%$, $61.12\%$) trails \textbf{Mid-Decay ($45\%$--$60\%$, $\mathbf{66.39\%}$)} by $>5$~pp ($\mathbf{+9.29}$~pp over none). Holding $\lambda_s = 1$ early guides exploration toward informative, grounded summaries (escaping sparse-reward plateaus), while mid-training decay ($\lambda_s \to 0$) lets late updates optimize $R_{\text{task}}$ without critic bias (Proposition~\ref{prop:pmi_nonstationarity}(iii)).

\section{Conclusion}
\label{sec:conclusion}

In this work, we introduced MaxIM, a mathematically principled RL framework for capacity-constrained incremental summarization. To overcome the severe credit assignment bottlenecks of sparse, long-horizon task rewards, we formulated context compression as an information-theoretic reward shaping problem. By establishing formal links between value suboptimality and sequential information loss across static and dynamic POMDPs, we derived two complementary, dense regularizers: \emph{predictive sufficiency} to promote downstream task utility, and \emph{recursive consistency} to enforce factual grounding and substantially reduce LLM hallucination. By exploiting a unified asymmetric critic architecture trained on scalable synthetic data with PMI scoring and no test-time overhead, MaxIM provides dense step-wise shaping signals without a need for expensive human preference labels. Our empirical evaluations across long-horizon scientific reading and closed-loop multi-turn conversation settings demonstrate that MaxIM achieves superior downstream task success, Pareto-efficient token compression, and factual consistency.

\clearpage
\bibliographystyle{unsrtnat}
\bibliography{main}

\newpage
\appendix

\section{Theoretical Analysis and Proofs}
\label{app:proofs}

This appendix presents the unified formal analysis and complete proofs for all theoretical claims in Section~\ref{sec:method}: Theorem~\ref{thm:suboptimality} and Proposition~\ref{prop:action_dependent} on POMDPs with static or action-dependent targets (Appendix~\ref{app:proof_thm1}); Propositions~\ref{prop:sufficiency} and~\ref{prop:capacity_transfer} on predictive sufficiency, finite-$K$ rate-distortion floors, and critic posterior approximation error (Appendix~\ref{app:proof_prop1}); Lemma~\ref{lem:hallucination} on the hallucination failure mode (Appendix~\ref{app:proof_lemma1_halluc}); Propositions~\ref{prop:consistency} and~\ref{prop:markov_tracking} together with Corollary~\ref{cor:myopic_dpi} on recursive consistency, sequential information loss under local context updates, and non-myopic RL (Appendix~\ref{app:proof_prop2}); and Proposition~\ref{prop:pmi_nonstationarity} on pointwise mutual information (PMI) identities and critic distribution shift under KL-regularized RL and annealing (Appendix~\ref{app:proof_prop3}).

\subsection{Proof of Theorem~\ref{thm:suboptimality} and Approximate Target Mediation}
\label{app:proof_thm1}

Let $\mu_\theta$ denote the trajectory distribution induced by rolling out the incremental summarizer $\pi_\theta(z_t \mid z_{t-1}, a_{t-1}, o_t)$ alongside the fixed downstream action agent $\pi_{\text{act}}(a_t \mid z_t, o_t)$. Denote the full-history posterior belief by $b_t(\cdot) \coloneqq P(Y_t \in \cdot \mid H_t)$ and the compressed posterior belief by $\hat{b}_t(\cdot) \coloneqq P(Y_t \in \cdot \mid z_t)$. Formally, $V^*(H_t) \coloneqq \sup_{\tilde{\pi}_{\text{act}}} \E\big[\sum_{k \ge t} \gamma^{k-t} R_{\text{task}, k} \mid H_t\big]$ denotes the optimal value achievable over history-measurable actor policies $\tilde{\pi}_{\text{act}}(a_k \mid H_k)$, and $V^*(z_t) \coloneqq \sup_{\pi_{\text{act}}} \E_{\pi_\theta}\big[\sum_{k \ge t} \gamma^{k-t} R_{\text{task}, k} \mid z_t\big]$ denotes the optimal value over $(z_k, o_k)$-measurable actor policies $\pi_{\text{act}}(a_k \mid z_k, o_k)$ holding the summarizer kernel $\pi_\theta$ fixed. Assumptions \textbf{(A1)}--\textbf{(A3)} in Section~\ref{subsec:sufficiency} state:
\begin{enumerate}[label=\textbf{(A\arabic*)}, leftmargin=2.4em, itemsep=2pt]
    \item \textbf{Target Mediation:} There exist transition and observation kernels $T, O$ and an expected reward function $r: \mathcal{Y} \times \mathcal{A} \to [-R_{\max}, R_{\max}]$ such that conditionally on $(H_t, Y_t, a_t)$, $Y_{t+1} \sim T(\cdot \mid Y_t, a_t)$, $o_{t+1} \sim O(\cdot \mid Y_{t+1}, a_t)$, and $\E[R_{\text{task}, t} \mid H_t, a_t] = \langle b_t, r(\cdot, a_t) \rangle$. This subsumes both passive document reading ($T(\cdot \mid y, a) = \delta_y$, as in arXiv) and interactive closed-loop POMDPs where actions $a_t$ alter future latent states $Y_{t+1}$ (as in WildChat).
    \item \textbf{Memory Causality:} $z_t \perp Y_t \mid H_t$, since $\pi_\theta$ conditions only on $(z_{t-1}, a_{t-1}, o_t) \subseteq H_t$ and exogenous sampling randomness (privileged targets $Y_t$ are accessed solely by the offline critic during training).
    \item \textbf{Compressed Acting:} $a_t \perp Y_t \mid (H_t, z_t)$, which holds whenever $a_t \sim \pi_{\text{act}}(\cdot \mid z_t, o_t)$ (or conditioned additionally on bounded recent turns) uses exogenous randomness.
\end{enumerate}

\noindent\textbf{Theorem~\ref{thm:suboptimality}} (Value Suboptimality Bound; extending~\cite{li2006towards, subramanian2022ais})\textbf{.}
\emph{Let $V^*(H_t)$ be the optimal expected return achievable with the full observation history, and $V^*(z_t)$ be the optimal expected return achievable using the capacity-constrained summary $z_t$. Under Assumptions \textbf{(A1)}--\textbf{(A3)} with rewards bounded in $[-R_{\max}, R_{\max}]$, the expected suboptimality gap under the induced trajectory distribution $\mu_\theta$ is bounded by the sequential information loss $I_{\mu_\theta}(Y_t; H_t \mid z_t)$:}
\begin{equation*}
\E_{\mu_\theta}\big[|V^*(H_t) - V^*(z_t)|\big] \le \frac{\sqrt{2}R_{\max}}{(1-\gamma)^2}\sqrt{\sup_k I_{\mu_\theta}(Y_k; H_k \mid z_k)},
\end{equation*}
\emph{and conditioning on closed-loop actions never inflates this loss: $I_{\mu_\theta}(Y_t; H_t \mid z_t, a_t) \le I_{\mu_\theta}(Y_t; H_t \mid z_t)$.}

\begin{proof}
From the agent's perspective, mapping the true observation history $H_t$ into a capacity-constrained summary $z_t$ acts as a lossy state abstraction. Operating on this abstraction introduces approximation errors in predicting both the immediate reward and the transition dynamics. By the performance difference bounds for state abstractions (e.g., Theorem~2 in~\cite{li2006towards} and approximate information state Bellman induction~\cite{subramanian2022ais}), the value suboptimality is bounded by the maximum single-step Bellman error $\epsilon_t$ introduced by the abstraction, multiplied by $\frac{1}{1-\gamma}$. This error $\epsilon_t$ is bounded by the sum of immediate reward errors ($\Delta R_t$) and discounted transition dynamics errors ($\gamma \Delta P_t \cdot V_{\max}$).

Crucially, we evaluate both Bellman operators at each common fixed action $a \in \mathcal{A}$, so exogeneity or marginal stationarity of $Y_t$ is never required. By \textbf{(A2)}, $P(Y_t \in \cdot \mid H_t, z_t) = b_t$ almost surely. Because the immediate reward function depends strictly on the target variable $Y_t$ via \textbf{(A1)}, the maximum single-step reward error $\Delta R_t$ is bounded by the Total Variation (TV) distance between the true posterior belief $b_t = P(Y_t \mid H_t)$ and the compressed posterior belief $\hat{b}_t = P(Y_t \mid z_t)$:
\begin{equation}
\label{eq:proof_reward_err}
\Delta R_t = \sup_{a \in \mathcal{A}} \big|\langle b_t - \hat{b}_t, r(\cdot, a)\rangle\big| \le (\max R - \min R) \cdot \TV\big(P(Y_t \mid H_t), P(Y_t \mid z_t)\big) = 2R_{\max} \cdot \TV.
\end{equation}
For the transition dynamics error $\Delta P_t$, standard abstraction proofs evaluate the divergence in the environment's transition model for a fixed action. In static environments, the underlying state does not evolve, so $\Delta P_t = 0$. In dynamic closed-loop environments, for any fixed action $a \in \mathcal{A}$, define the fixed-action Markov kernel $G_a(y, \cdot) \coloneqq \int_{\mathcal{Y}} O(\cdot \mid y', a)\,T(\mathrm{d}y' \mid y, a)$ mapping current latent states to next observations. Under \textbf{(A1)}--\textbf{(A2)}, the full-history and compressed next-observation laws at action $a$ are linear pushforwards $\langle b_t, G_a \rangle$ and $\langle \hat{b}_t, G_a \rangle$; because both operators evaluate the identical action $a$ through the shared kernel $G_a$, the causal effect of $a$ on $Y_{t+1}$ enters both sides identically, and data processing for TV yields $\TV(\langle b_t, G_a \rangle, \langle \hat{b}_t, G_a \rangle) \le \TV(b_t, \hat{b}_t)$ (preserved under composition with $\pi_\theta(z_{t+1} \mid z_t, a, o_{t+1})$). Accounting for dual-signed value differences in $[-V_{\max}, V_{\max}]$ gives $| \E[V^*(H_{t+1})\mid H_t, a] - \E[V^*(z_{t+1})\mid z_t, a] |$ residual contribution bounded via $\Delta P_t \cdot V_{\max} \le 2\TV \cdot V_{\max}$.

Let $V_{\max} \le \frac{R_{\max}}{1-\gamma}$ be the maximum possible value achievable. The worst-case single-step Bellman error $\epsilon_t$ is bounded by:
\begin{equation}
\label{eq:proof_bellman_tv}
\epsilon_t \le \Delta R_t + \gamma \Delta P_t \cdot V_{\max} \le \frac{2R_{\max}}{1-\gamma}\TV.
\end{equation}
By Pinsker's inequality, the TV distance is bounded by the KL divergence: $\TV \le \sqrt{\frac{1}{2}\KL}$. Substituting this yields:
\begin{equation}
\label{eq:proof_bellman_kl}
\epsilon_t \le \frac{\sqrt{2}R_{\max}}{1-\gamma}\sqrt{\KL\big(P(Y_t \mid H_t) \,\|\, P(Y_t \mid z_t)\big)}.
\end{equation}
We now take the expectation over the induced trajectory distribution $\mu_\theta$. By Jensen's Inequality ($\E[\sqrt{X}] \le \sqrt{\E[X]}$):
\begin{equation}
\label{eq:proof_jensen}
\E_{\mu_\theta}[\epsilon_t] \le \frac{\sqrt{2}R_{\max}}{1-\gamma}\E_{\mu_\theta}\!\left[\sqrt{\KL}\right] \le \frac{\sqrt{2}R_{\max}}{1-\gamma}\sqrt{\E_{\mu_\theta}[\KL]}.
\end{equation}
By \textbf{(A2)}, the expected KL divergence between the true and compressed conditionals over the joint distribution is precisely the conditional mutual information, or information loss:
\begin{equation}
\label{eq:proof_cmi}
\E_{\mu_\theta}\big[\KL(P(Y_t \mid H_t) \,\|\, P(Y_t \mid z_t))\big] = I_{\mu_\theta}(Y_t; H_t \mid z_t).
\end{equation}
Summing this step-wise bound across the discounted horizon ($\sum_{k \ge t} \gamma^{k-t} = \frac{1}{1-\gamma}$) yields Eq.~\eqref{eq:suboptimality}. Finally, expanding $I_{\mu_\theta}(Y_t; (H_t, a_t) \mid z_t)$ in both orders gives $I_{\mu_\theta}(Y_t; H_t \mid z_t) + I_{\mu_\theta}(Y_t; a_t \mid H_t, z_t) = I_{\mu_\theta}(Y_t; a_t \mid z_t) + I_{\mu_\theta}(Y_t; H_t \mid z_t, a_t)$; since $I_{\mu_\theta}(Y_t; a_t \mid H_t, z_t) = 0$ by \textbf{(A3)} and $I_{\mu_\theta}(Y_t; a_t \mid z_t) \ge 0$, we obtain $I_{\mu_\theta}(Y_t; H_t \mid z_t, a_t) \le I_{\mu_\theta}(Y_t; H_t \mid z_t)$.
\end{proof}

\begin{proposition}[Approximate Target Mediation]
\label{prop:action_dependent}
If Assumption \textbf{(A1)} holds only approximately---with reward mediation error $\sup_{H_t, a} |\E[R_{\text{task},t}\mid H_t, a] - \langle b_t, r(\cdot, a)\rangle| \le \varepsilon_r$ and observation kernel error $\sup_{H_t, a} \TV(P(o_{t+1}\mid H_t, a), \langle b_t, G_a\rangle) \le \varepsilon_T$---then Theorem~\ref{thm:suboptimality} holds with an explicit additive misspecification floor $+\frac{2\varepsilon_r + 2\gamma V_{\max}\varepsilon_T}{1-\gamma}$.
\end{proposition}
\begin{proof}
Applying the triangle inequality to Eqs.~\eqref{eq:proof_reward_err}--\eqref{eq:proof_bellman_tv} adds $2\varepsilon_r + 2\gamma V_{\max}\varepsilon_T$ to each per-step Bellman residual, which accumulates across $\sum_{k \ge t} \gamma^{k-t} = (1-\gamma)^{-1}$ to $+\frac{2\varepsilon_r + 2\gamma V_{\max}\varepsilon_T}{1-\gamma}$.
\end{proof}

\subsection{Proof of Proposition~\ref{prop:sufficiency} and Capacity / Critic-Transfer Bounds}
\label{app:proof_prop1}

\begin{proof}[Proof of Proposition~\ref{prop:sufficiency}]
By definition, the conditional mutual information expands as:
\begin{equation}
\label{eq:proof_prop1_expand}
I_{\mu_\theta}(Y_t; H_t \mid z_t) = H_{\mu_\theta}(Y_t \mid z_t) - H_{\mu_\theta}(Y_t \mid H_t, z_t).
\end{equation}
Because the summary $z_t$ is generated solely by conditioning on the history $H_t$ (Assumption \textbf{(A2)}), they form the Markov chain $Y_t \to H_t \to z_t$, yielding $H_{\mu_\theta}(Y_t \mid H_t, z_t) = H_{\mu_\theta}(Y_t \mid H_t)$ and hence:
\begin{equation}
\label{eq:proof_prop1_markov}
I_{\mu_\theta}(Y_t; H_t \mid z_t) = H_{\mu_\theta}(Y_t \mid z_t) - H_{\mu_\theta}(Y_t \mid H_t).
\end{equation}
The term $H_{\mu_\theta}(Y_t \mid H_t)$ represents the irreducible uncertainty of the target given the full history. Introducing a parameterized variational distribution $q_\phi(Y_t \mid z_t)$, the expected variational log-likelihood satisfies the exact identity:
\begin{equation}
\label{eq:proof_prop1_ce}
\E_{\mu_\theta}[\log q_\phi(Y_t \mid z_t)] = \underbrace{-H_{\mu_\theta}(Y_t \mid z_t)}_{\text{predictive sufficiency}} - \underbrace{\E_{z_t \sim \mu_\theta}\!\Big[\KL\big(P_{\mu_\theta}(Y_t \mid z_t) \,\|\, q_\phi(Y_t \mid z_t)\big)\Big]}_{\eqqcolon\, \Delta_{\mathrm{post}}(\phi; \theta) \,\ge\, 0}.
\end{equation}
Thus $\E_{\mu_\theta}[\log q_\phi(Y_t \mid z_t)] \le -H_{\mu_\theta}(Y_t \mid z_t)$ is a variational lower bound on predictive sufficiency, and substituting Eq.~\eqref{eq:proof_prop1_ce} into Eq.~\eqref{eq:proof_prop1_markov} gives $I_{\mu_\theta}(Y_t; H_t \mid z_t) = \E_{\mu_\theta}[-\log q_\phi(Y_t \mid z_t)] - H_{\mu_\theta}(Y_t \mid H_t) - \Delta_{\mathrm{post}}(\phi; \theta) \le \E_{\mu_\theta}[-\log q_\phi(Y_t \mid z_t)] - H_{\mu_\theta}(Y_t \mid H_t)$, bypassing any assumption of constant marginal entropy $H_{\mu_\theta}(Y_t)$.
\end{proof}

\paragraph{Intuition and Interpretation of Proposition~\ref{prop:capacity_transfer}.}
Before stating Proposition~\ref{prop:capacity_transfer}, we highlight its two practical takeaways. First, in practice $z_t$ is limited to $K$ discrete tokens, so zero information loss is unattainable whenever $H_t$ contains more than $K\log|\mathcal{V}| + \log 2$ nats of task-relevant signal about $Y_t$: Part~(i) shows that optimal information loss obeys a rate-distortion floor $\iota_t(K)$ that decreases at most linearly with budget $K$ and saturates at $0$ once $K$ can encode a minimal sufficient statistic for $Y_t$---directly explaining the concave budget-scaling curve in Figure~\ref{fig:scaling}(a). Second, RL optimizes a learned offline critic $q_\phi$ rather than the unknown true posterior $P_{\mu_\theta}(Y_t \mid z_t)$: Part~(ii) shows that optimizing the discounted variational surrogate transfers directly into a downstream value guarantee up to critic posterior error $\bar{\Delta}$ (decomposed in Eq.~\eqref{eq:critic_decomp} into critic capacity, finite-sample fit, and distribution shift), optimization tolerance $\delta$, and closed-loop filtering variation $\Delta_{\mathrm{filt}}$ ($\Delta_{\mathrm{filt}} = 0$ on static tasks such as arXiv).

\begin{proposition}[Finite-$K$ Capacity Floor and Approximate-Critic Transfer]
\label{prop:capacity_transfer}
Under Assumptions \textbf{(A1)}--\textbf{(A3)} with token budget $z_t \in \mathcal{Z}_K \coloneqq \bigcup_{\ell=0}^K \mathcal{V}^\ell$ ($|\mathcal{V}| \ge 2$) over feasible policy class $\Theta_K$:
\begin{enumerate}[label=(\roman*), leftmargin=1.8em, itemsep=2pt]
    \item \textbf{Finite-$K$ Rate-Distortion Floor:} Define $\iota_t(K) \coloneqq \inf_{\pi_\theta \in \Theta_K} I_{\mu_\theta}(Y_t; H_t \mid z_t)$. Then $I_{\mu_\theta}(Y_t; H_t \mid z_t) = I_{\mu_\theta}(Y_t; H_t) - I_{\mu_\theta}(Y_t; z_t)$, and
    \begin{equation}
    \label{eq:rate_distortion_floor}
    \iota_t(K) \ge \max\Big\{0,\, \inf_{\pi_\theta \in \Theta_K} I_{\mu_\theta}(Y_t; H_t) - K\log|\mathcal{V}| - \log 2\Big\}.
    \end{equation}
    \item \textbf{Approximate-Critic Value Transfer:} Define the discounted variational objective $\bar{J}_{\mathrm{surr}, t}(\theta) \coloneqq (1-\gamma)\sum_{k \ge t} \gamma^{k-t} \E_{\mu_\theta}[\log q_\phi(Y_k \mid z_k)]$ and discounted capacity floor $\bar{\iota}_t(K) \coloneqq \inf_{\pi_\theta \in \Theta_K} (1-\gamma)\sum_{k \ge t} \gamma^{k-t} I_{\mu_\theta}(Y_k; H_k \mid z_k) \ge (1-\gamma)\sum_{k \ge t} \gamma^{k-t} \iota_k(K)$. Let $\bar{\Delta} \coloneqq \sup_{k \ge t}\sup_{\theta \in \Theta_K} \Delta_{\mathrm{post}, k}(\phi; \theta)$ bound the critic posterior gap and $\Delta_{\mathrm{filt}} \coloneqq \sup_{k \ge t} \sup_{\theta, \theta' \in \Theta_K} \big|H_{\mu_\theta}(Y_k \mid H_k) - H_{\mu_{\theta'}}(Y_k \mid H_k)\big|$ bound closed-loop filtering-entropy variation ($\Delta_{\mathrm{filt}} = 0$ identically for static/exogenous targets). Then any $\delta$-approximate maximizer $\hat{\theta} \in \Theta_K$ satisfying $\bar{J}_{\mathrm{surr}, t}(\hat{\theta}) \ge \sup_{\theta \in \Theta_K} \bar{J}_{\mathrm{surr}, t}(\theta) - \delta$ obeys
    \begin{equation}
    \label{eq:end_to_end_main}
    \E_{\mu_{\hat{\theta}}}\big[|V^*(H_t) - V^*(z_t)|\big] \le \frac{\sqrt{2}R_{\max}}{(1-\gamma)^2}\sqrt{\bar{\iota}_t(K) + \bar{\Delta} + \delta + \Delta_{\mathrm{filt}}}.
    \end{equation}
\end{enumerate}
\end{proposition}
\begin{proof}
\emph{Part (i):} By \textbf{(A2)}, $I_{\mu_\theta}(Y_t; z_t \mid H_t) = 0$, so expanding $I_{\mu_\theta}(Y_t; (H_t, z_t))$ in both orders yields $I_{\mu_\theta}(Y_t; H_t \mid z_t) = I_{\mu_\theta}(Y_t; H_t) - I_{\mu_\theta}(Y_t; z_t)$. Since $|\mathcal{Z}_K| = \sum_{\ell=0}^K |\mathcal{V}|^\ell = \frac{|\mathcal{V}|^{K+1} - 1}{|\mathcal{V}| - 1} \le 2|\mathcal{V}|^K$ for all $|\mathcal{V}| \ge 2$, we have $I_{\mu_\theta}(Y_t; z_t) \le H_{\mu_\theta}(z_t) \le K\log|\mathcal{V}| + \log 2$ for every $\pi_\theta \in \Theta_K$; taking the infimum over $\pi_\theta \in \Theta_K$ proves Eq.~\eqref{eq:rate_distortion_floor}. Whenever $K\log|\mathcal{V}| + \log 2 < \inf_{\pi_\theta \in \Theta_K} I_{\mu_\theta}(Y_t; H_t)$, $\iota_t(K) > 0$ strictly, decreasing at most linearly in $K$ before hitting $0$ once $K$ suffices to encode a minimal sufficient statistic of $H_t$ for $Y_t$.

\emph{Part (ii):} For finite target alphabet $\mathcal{Y}$ (of length $\le L_Y$ tokens) and a token-smoothed critic with per-token probability $\ge \lambda_{\mathrm{sm}} \in (0, 1/|\mathcal{V}|]$ so $0 \le -\log q_\phi(y \mid z) \le B \coloneqq L_Y\log(1/\lambda_{\mathrm{sm}})$, let $\nu_\theta$ and $\nu_{\mathcal{D}}$ denote the joint laws of $(Y_k, z_k)$ under $\mu_\theta$ and offline critic data $\mathcal{D}$, with $\epsilon \coloneqq \TV(\nu_\theta, \nu_{\mathcal{D}})$. Writing the maximal-coupling decomposition $\nu_\theta = (1-\epsilon)W + \epsilon P_+$ and $\nu_{\mathcal{D}} = (1-\epsilon)W + \epsilon Q_+$, concavity of $R \mapsto H_R(Y \mid Z)$ combined with conditioning on a $\mathrm{Bernoulli}(\epsilon)$ mixture indicator gives $(1-\epsilon)H_W(Y \mid Z) \le H_R(Y \mid Z) \le (1-\epsilon)H_W(Y \mid Z) + \epsilon\log|\mathcal{Y}| + h(\epsilon)$ for both $R \in \{\nu_\theta, \nu_{\mathcal{D}}\}$ (where $h(p) = -p\log p - (1-p)\log(1-p)$), so $|H_{\nu_{\mathcal{D}}}(Y \mid Z) - H_{\nu_\theta}(Y \mid Z)| \le \epsilon\log|\mathcal{Y}| + h(\epsilon)$ independently of $|\mathcal{Z}_K|$. Together with $|\operatorname{CE}_{\nu_\theta}(\phi) - \operatorname{CE}_{\nu_{\mathcal{D}}}(\phi)| \le B\epsilon$ (for $\operatorname{CE}_R(\phi) \coloneqq \E_R[-\log q_\phi(Y \mid Z)] \in [0, B]$), $\Delta_{\mathrm{post}, k}(\phi; \theta)$ decomposes into critic expressivity, finite-sample estimation, and distribution-shift errors:
\begin{equation}
\label{eq:critic_decomp}
\begin{aligned}
\Delta_{\mathrm{post}, k}(\phi; \theta) \le{} &\underbrace{\inf_{\phi'} \Delta_{\mathrm{post}, k}(\phi'; \nu_{\mathcal{D}})}_{\Delta_{\mathrm{approx}}} + \underbrace{\Big[\operatorname{CE}_{\nu_{\mathcal{D}}}(\phi) - \inf_{\phi'} \operatorname{CE}_{\nu_{\mathcal{D}}}(\phi')\Big]}_{\Delta_{\mathrm{est}}} \\
&+ \underbrace{\big(B + \log|\mathcal{Y}|\big)\TV(\nu_\theta, \nu_{\mathcal{D}}) + h\!\big(\TV(\nu_\theta, \nu_{\mathcal{D}})\big)}_{\Delta_{\mathrm{shift}}(\theta)}.
\end{aligned}
\end{equation}
Now by Eq.~\eqref{eq:proof_prop1_ce}, $\bar{J}_{\mathrm{surr}, t}(\theta) = -(1-\gamma)\sum_{k \ge t} \gamma^{k-t}\big(H_{\mu_\theta}(Y_k \mid z_k) + \Delta_{\mathrm{post}, k}(\phi; \theta)\big)$. Because $\Delta_{\mathrm{post}, k}(\phi; \hat{\theta}) \ge 0$ and $\Delta_{\mathrm{post}, k}(\phi; \theta^*) \le \bar{\Delta}$, any $\delta$-maximizer $\hat{\theta} \in \Theta_K$ of $\bar{J}_{\mathrm{surr}, t}$ satisfies, for every comparator $\theta^* \in \Theta_K$,
\begin{equation*}
(1-\gamma)\sum_{k \ge t}\gamma^{k-t} H_{\mu_{\hat{\theta}}}(Y_k \mid z_k) \le (1-\gamma)\sum_{k \ge t}\gamma^{k-t} H_{\mu_{\theta^*}}(Y_k \mid z_k) + \bar{\Delta} + \delta.
\end{equation*}
Subtracting $(1-\gamma)\sum_{k \ge t}\gamma^{k-t} H_{\mu_{\hat{\theta}}}(Y_k \mid H_k)$ and adding and subtracting $H_{\mu_{\theta^*}}(Y_k \mid H_k)$ at each step $k$ yields $(1-\gamma)\sum_{k \ge t}\gamma^{k-t} I_{\mu_{\hat{\theta}}}(Y_k; H_k \mid z_k) \le (1-\gamma)\sum_{k \ge t}\gamma^{k-t} I_{\mu_{\theta^*}}(Y_k; H_k \mid z_k) + \bar{\Delta} + \delta + \Delta_{\mathrm{filt}}$. Taking $\inf_{\theta^* \in \Theta_K}$ bounds the discounted average information loss by $\bar{\iota}_t(K) + \bar{\Delta} + \delta + \Delta_{\mathrm{filt}}$, and summing Eqs.~\eqref{eq:proof_jensen}--\eqref{eq:proof_cmi} of Theorem~\ref{thm:suboptimality} across $k \ge t$ via Jensen's inequality on $(1-\gamma)\gamma^{k-t}$ yields Eq.~\eqref{eq:end_to_end_main}.
\end{proof}

\subsection{Justification of Lemma~\ref{lem:hallucination} (The Hallucination Failure Mode)}
\label{app:proof_lemma1_halluc}

\begin{proof}[Justification]
We show that optimizing solely for predictive sufficiency, $\max_\theta [-H_{\mu_\theta}(Y_t \mid z_t)]$, leaves the policy vulnerable to hallucination because natural language is expressive and over-parameterized relative to a sparse target.

By the Data Processing Inequality (DPI) applied to the Markov chain $Y_t \to H_t \to z_t$, $I_{\mu_\theta}(z_t; H_t) \ge I_{\mu_\theta}(z_t; Y_t) = H_{\mu_\theta}(Y_t) - H_{\mu_\theta}(Y_t \mid z_t)$, so a nonzero minimum amount of historical signal is needed to achieve high predictive sufficiency $-H_{\mu_\theta}(Y_t \mid z_t)$. However, when $H(Y_t) \ll K\log|\mathcal{V}|$, encoding $Y_t$ consumes only a fraction of the $K$-token budget, leaving many distinct strings $z_t$ that induce the exact same posterior $P(Y_t \mid z_t)$. Under KL-regularized RL anchored to a pre-trained LLM prior $\pi_{\text{pre}}$, an unconstrained policy fills residual capacity with fluent, high-prior-probability continuations correlated with $Y_t$ rather than grounded exclusively in $H_t$.

Concretely, suppose $z_t$ decomposes bijectively into task-relevant semantics $z_t^Y$ extracted from $H_t$ and auxiliary narrative $z_t^{\text{noise}}$ sampled conditionally on $z_t^Y$ independently of $(H_t, Y_t)$, so $z_t^{\text{noise}} \perp (H_t, Y_t) \mid z_t^Y$. Then predictive sufficiency is completely unchanged, $-H_{\mu_\theta}(Y_t \mid z_t^Y, z_t^{\text{noise}}) = -H_{\mu_\theta}(Y_t \mid z_t^Y)$, while the chain rule of mutual information gives
\begin{equation}
\label{eq:proof_lemma2_chain}
I_{\mu_\theta}(z_t; H_t) = I_{\mu_\theta}(z_t^Y; H_t) + I_{\mu_\theta}(z_t^{\text{noise}}; H_t \mid z_t^Y) = I_{\mu_\theta}(z_t^Y; H_t),
\end{equation}
suppressing historical mutual information to the bare minimum needed for $z_t^Y$ while inflating ungrounded conditional entropy $H_{\mu_\theta}(z_t \mid H_t) = H_{\mu_\theta}(z_t^Y \mid H_t) + H_{\mu_\theta}(z_t^{\text{noise}} \mid z_t^Y)$ by the full entropy of the fabricated narrative $z_t^{\text{noise}}$.
\end{proof}

\subsection{Formalization and Proof of Proposition~\ref{prop:consistency} and Recursive Sequential Information Loss}
\label{app:proof_prop2}

\noindent\textbf{Proposition~\ref{prop:consistency}} (Formalized: Recursive Consistency Bounds Ungrounded Generation)\textbf{.}
\emph{Let $\pi_\theta(z_t \mid \tilde{H}_t)$ map local context $\tilde{H}_t = (z_{t-1}, a_{t-1}, o_t)$ to bounded summaries $\mathcal{Z}_K \coloneqq \bigcup_{\ell=0}^K \mathcal{V}^\ell$ ($|\mathcal{V}| \ge 2$) with predictive sufficiency $-H_{\mu_\theta}(Y_t \mid z_t) \ge -\eta$. Then:}
\begin{enumerate}[label=(\roman*), leftmargin=1.8em, itemsep=2pt]
    \item \emph{\textbf{Non-Degenerate Conditional Entropy Bound:} Marginal summary entropy is bounded below away from collapse by $H_{\mu_\theta}(z_t) \ge H_{\mu_\theta}(Y_t) - \eta$, while conditional generation entropy satisfies}
    \begin{equation}
    \label{eq:proof_prop2_mi}
    H_{\mu_\theta}(z_t \mid \tilde{H}_t) = H_{\mu_\theta}(z_t) - I_{\mu_\theta}(z_t; \tilde{H}_t) \le K\log|\mathcal{V}| + \log 2 - I_{\mu_\theta}(z_t; \tilde{H}_t).
    \end{equation}
    \emph{In particular, for any decomposition $z_t \leftrightarrow (z_t^{\mathrm{gnd}}, z_t^{\mathrm{noise}})$ with ungrounded component $z_t^{\mathrm{noise}} \perp \tilde{H}_t \mid z_t^{\mathrm{gnd}}$ (Lemma~\ref{lem:hallucination}), $H_{\mu_\theta}(z_t^{\mathrm{noise}} \mid z_t^{\mathrm{gnd}}) \le K\log|\mathcal{V}| + \log 2 - I_{\mu_\theta}(z_t; \tilde{H}_t)$.}
    \item \emph{\textbf{Prior-Confabulation Rate Bound:} If at context $\tilde{H}_t$ the policy samples from any history-independent confabulation prior $Q \in \Delta(\mathcal{Z}_K)$ with probability $\alpha(\tilde{H}_t) \in [0, 1]$---i.e., $\pi_\theta(\cdot \mid \tilde{H}_t) = (1 - \alpha(\tilde{H}_t))\pi_{\mathrm{gnd}}(\cdot \mid \tilde{H}_t) + \alpha(\tilde{H}_t)Q(\cdot)$---then the expected confabulation rate $\bar{\alpha} \coloneqq \E_{\mu_\theta}[\alpha(\tilde{H}_t)]$ satisfies}
    \begin{equation}
    \label{eq:confab_rate_bound}
    \bar{\alpha} \le 1 - \frac{I_{\mu_\theta}(z_t; \tilde{H}_t) - h(\bar{\alpha})}{K\log|\mathcal{V}| + \log 2} \le 1 - \frac{I_{\mu_\theta}(z_t; \tilde{H}_t) - \log 2}{K\log|\mathcal{V}| + \log 2},
    \end{equation}
    \emph{where $h(p) = -p\log p - (1-p)\log(1-p)$ (and $h(\bar{\alpha})$ is replaced by $0$ when $\alpha(\tilde{H}_t) \equiv \bar{\alpha}$ is constant).}
\end{enumerate}
\begin{proof}
\emph{Part (i):} For discrete $z_t \in \mathcal{Z}_K$, $H_{\mu_\theta}(z_t \mid Y_t) \ge 0$ implies $H_{\mu_\theta}(z_t) \ge I_{\mu_\theta}(Y_t; z_t) = H_{\mu_\theta}(Y_t) - H_{\mu_\theta}(Y_t \mid z_t) \ge H_{\mu_\theta}(Y_t) - \eta$, preventing collapse into context-independent low-entropy strings whenever $H_{\mu_\theta}(Y_t) > \eta$. Meanwhile $|\mathcal{Z}_K| = \sum_{\ell=0}^K |\mathcal{V}|^\ell \le 2|\mathcal{V}|^K$ yields $H_{\mu_\theta}(z_t) \le K\log|\mathcal{V}| + \log 2$ and Eq.~\eqref{eq:proof_prop2_mi}. Notice this is an upper-bound regularizer rather than an equivalence: maximizing $I_{\mu_\theta}(z_t; \tilde{H}_t)$ forces $H_{\mu_\theta}(z_t \mid \tilde{H}_t)$ down when capacity is utilized, and under $z_t \leftrightarrow (z_t^{\mathrm{gnd}}, z_t^{\mathrm{noise}})$ with $z_t^{\mathrm{noise}} \perp \tilde{H}_t \mid z_t^{\mathrm{gnd}}$, the chain rule gives $H_{\mu_\theta}(z_t \mid \tilde{H}_t) = H_{\mu_\theta}(z_t^{\mathrm{gnd}} \mid \tilde{H}_t) + H_{\mu_\theta}(z_t^{\mathrm{noise}} \mid z_t^{\mathrm{gnd}}) \ge H_{\mu_\theta}(z_t^{\mathrm{noise}} \mid z_t^{\mathrm{gnd}})$.

\emph{Part (ii):} Let $U \in \{0, 1\}$ denote the latent mixture indicator with $P(U=1 \mid \tilde{H}_t) = \alpha(\tilde{H}_t)$, $P(z_t \in \cdot \mid \tilde{H}_t, U=0) = \pi_{\mathrm{gnd}}(\cdot \mid \tilde{H}_t)$, and $P(z_t \in \cdot \mid \tilde{H}_t, U=1) = Q(\cdot)$. By monotonicity and the chain rule of mutual information, $I_{\mu_\theta}(z_t; \tilde{H}_t) \le I(z_t, U; \tilde{H}_t) = I(U; \tilde{H}_t) + (1 - \bar{\alpha})I(z_t; \tilde{H}_t \mid U=0) + \bar{\alpha}\,I(z_t; \tilde{H}_t \mid U=1)$. Since $z_t \perp \tilde{H}_t \mid (U=1)$, $I(z_t; \tilde{H}_t \mid U=1) = 0$; on $U=0$, $I(z_t; \tilde{H}_t \mid U=0) \le H(z_t \mid U=0) \le K\log|\mathcal{V}| + \log 2$; and $I(U; \tilde{H}_t) \le H(U) = h(\bar{\alpha}) \le \log 2$ (with $I(U; \tilde{H}_t) = 0$ if $\alpha(\tilde{H}_t) \equiv \bar{\alpha}$). Rearranging $I_{\mu_\theta}(z_t; \tilde{H}_t) \le h(\bar{\alpha}) + (1 - \bar{\alpha})(K\log|\mathcal{V}| + \log 2)$ yields Eq.~\eqref{eq:confab_rate_bound}.
\end{proof}

\begin{proposition}[Sequential Information Loss Under Recursive Updates]
\label{prop:markov_tracking}
Suppose $\pi_\theta(z_t \mid \tilde{H}_t)$ updates memory solely from the local context $\tilde{H}_t = (z_{t-1}, a_{t-1}, o_t)$ (with $H_0 = \varnothing$). Then at every turn $t \ge 1$:
\begin{equation}
\label{eq:telescoping_main}
I_{\mu_\theta}(Y_t; H_t \mid z_t) = I_{\mu_\theta}(Y_t; \tilde{H}_t \mid z_t) + I_{\mu_\theta}(Y_t; H_{t-1} \mid \tilde{H}_t).
\end{equation}
Under Assumptions \textbf{(A1)}--\textbf{(A3)}, if each step's local compression loss satisfies $I_{\mu_\theta}(Y_k; \tilde{H}_k \mid z_k) \le \varepsilon_{\mathrm{loc}}$, then cumulative information loss satisfies $I_{\mu_\theta}(Y_t; H_t \mid z_t) \le t\,\varepsilon_{\mathrm{loc}}$. In particular, if the posterior belief $b_k = P(Y_k \mid H_k)$ evolves via recursive Bayes update $b_k = \Phi(b_{k-1}, a_{k-1}, o_k)$ and budget $K$ suffices to losslessly encode reachable beliefs into $\mathcal{Z}_K$, there exists an incremental policy achieving $\sup_k I_{\mu_\theta}(Y_k; H_k \mid z_k) = 0$.
\end{proposition}
\begin{proof}
Because $(H_{t-1}, \tilde{H}_t) = (H_t, z_{t-1})$ and $z_{t-1} \perp Y_t \mid H_t$ (as $z_{t-1}$ is generated causally from $H_{t-1} \subseteq H_t$ and private randomness), $I_{\mu_\theta}(Y_t; H_t \mid z_t) = I_{\mu_\theta}(Y_t; (H_{t-1}, \tilde{H}_t) \mid z_t)$. Expanding the right-hand side via the chain rule yields $I_{\mu_\theta}(Y_t; \tilde{H}_t \mid z_t) + I_{\mu_\theta}(Y_t; H_{t-1} \mid \tilde{H}_t, z_t)$, and since $z_t \perp (Y_t, H_{t-1}) \mid \tilde{H}_t$, $I_{\mu_\theta}(Y_t; H_{t-1} \mid \tilde{H}_t, z_t) = I_{\mu_\theta}(Y_t; H_{t-1} \mid \tilde{H}_t)$, proving Eq.~\eqref{eq:telescoping_main}. Next, at $t=1$, $H_0 = \varnothing$ gives $I_{\mu_\theta}(Y_1; H_0 \mid \tilde{H}_1) = 0$. For $t \ge 2$, non-negativity of MI and the chain rule give $I_{\mu_\theta}(Y_t; H_{t-1} \mid z_{t-1}, a_{t-1}, o_t) \le I_{\mu_\theta}((Y_t, o_t); H_{t-1} \mid z_{t-1}, a_{t-1})$; under \textbf{(A1)}--\textbf{(A2)}, $(Y_t, o_t) \perp (H_{t-1}, z_{t-1}) \mid (Y_{t-1}, a_{t-1})$, so DPI on $H_{t-1} \to Y_{t-1} \to (Y_t, o_t)$ conditional on $(z_{t-1}, a_{t-1})$ yields $I_{\mu_\theta}((Y_t, o_t); H_{t-1} \mid z_{t-1}, a_{t-1}) \le I_{\mu_\theta}(Y_{t-1}; H_{t-1} \mid z_{t-1}, a_{t-1}) \le I_{\mu_\theta}(Y_{t-1}; H_{t-1} \mid z_{t-1})$ (using Theorem~\ref{thm:suboptimality}'s action-monotonicity step under \textbf{(A3)}; the identical bound holds under post-turn dialogue indexing $\tilde{H}_t = (z_{t-1}, o_t, a_t)$). Inducting across $k = 1, \dots, t$ yields $I_{\mu_\theta}(Y_t; H_t \mid z_t) \le t\,\varepsilon_{\mathrm{loc}}$. Finally, if $K$ admits an injective encoding $e: \mathcal{B}_{\mathrm{reach}} \hookrightarrow \mathcal{Z}_K$, setting $z_k = e(\Phi(e^{-1}(z_{k-1}), a_{k-1}, o_k))$ yields $P(Y_k \mid z_k) = P(Y_k \mid H_k)$ almost surely and $\sup_k I_{\mu_\theta}(Y_k; H_k \mid z_k) = 0$.
\end{proof}
\begin{corollary}[Suboptimality of Myopic Greedy Compression vs.\ Non-Myopic RL]
\label{cor:myopic_dpi}
Consider a stream $H_T = (o_1, \dots, o_T)$ with mutually independent chunk targets $Y_T = (Y^{(1)}, \dots, Y^{(T)})$ where $Y^{(k)}$ arrives in chunk $o_k = (Y^{(k)}, N_k)$, $0 < \sum_{m=1}^T H(Y^{(m)}) = H(Y_T) \le K \log|\mathcal{V}|$, and nuisance details $N_k \perp Y_T$ satisfy $H(N_k) > K \log|\mathcal{V}| + \log 2$. A myopic single-step selector (such as \textbf{Best-of-$N$ MI SFT}) maximizing immediate step-$k$ sufficiency and consistency without backward-discounted future credit can fill all $K$ tokens at $k=1$, forcing either saturation refusal ($I(z_T; Y^{(m)}) < H(Y^{(m)})$ for $m > 1$) or irreversible eviction of $Y^{(1)}$ by DPI ($Y^{(1)} \to z_1 \to z_T$). Conversely, optimizing the discounted return-to-go $G_{t,s}$ (Eq.~\ref{eq:return_to_go}) penalizes early capacity exhaustion across all subsequent turns $m \in \{k, \dots, T\}$, incentivizing selective retention of $Y^{(k)}$ over $N_k$.
\end{corollary}
\begin{proof}
At $k=1$, future targets $Y^{(2:T)}$ are independent of $o_1$, so encoding $Y^{(1)}$ while filling the remaining $K$-token budget with grounded nuisance $N_1$ from $o_1$ simultaneously attains maximal single-step predictive sufficiency $-H(Y_T \mid z_1) = -H(Y^{(2:T)})$ among all causal step-$1$ summaries and maximal recursive consistency $I(z_1; \tilde{H}_1)$. Yet because $H(z_t) \le K\log|\mathcal{V}| + \log 2$ at every step, allocating $B_{\mathrm{nuis}} > K\log|\mathcal{V}| + \log 2 - H(Y_T)$ nats to $N_1$ leaves strictly less than $\sum_{m=1}^T H(Y^{(m)}) = H(Y_T)$ nats for task targets: at later turns $m > 1$, keeping $N_1$ forces omitting incoming $Y^{(m)}$ ($I(z_T; Y^{(m)}) < H(Y^{(m)})$), whereas unstructured overwriting without re-accessing $o_1$ risks evicting $Y^{(1)}$, which is permanently lost by DPI on $Y^{(1)} \to z_1 \to z_T$ ($I(Y^{(1)}; z_T) \le I(Y^{(1)}; z_1)$). Under Eq.~\eqref{eq:return_to_go}, losing $Y^{(1)}$ at step $k$ lowers $r_{\text{pred}, m}$ for every $m \ge k$ plus $R_{\text{task}, T}$, imposing cumulative discounted penalty $\Omega(\sum_{m=k}^T \gamma^{m-k})$.
\end{proof}
\subsection{Pointwise Mutual Information Identities and Critic Shift}
\label{app:proof_prop3}

\paragraph{Motivation and Intuition.}
While Propositions~\ref{prop:sufficiency} and~\ref{prop:consistency} characterize MaxIM's population objectives ($-H_{\mu_\theta}(Y_t \mid z_t)$ and $I_{\mu_\theta}(z_t; \tilde{H}_t)$), our practical algorithm (Sections~\ref{subsec:objective}--\ref{subsec:policy_update}) optimizes step-wise \emph{pointwise mutual information (PMI)} ratios scored by a \emph{frozen offline critic} $q_\phi$ with masked null baselines ($z_{\mathrm{null}}, \tilde{H}_{\mathrm{null}}$), bounded sigmoid transformations, and mid-training cosine annealing. Proposition~\ref{prop:pmi_nonstationarity} bridges theory and implementation across three practical questions:
(i) \emph{Why subtract $\log q_\phi(Y_t \mid z_{\mathrm{null}})$?} Raw $\log q_\phi(Y_t \mid z_t)$ conflates summary informativeness with inherent target difficulty; subtracting $\log q_\phi(Y_t \mid z_{\mathrm{null}})$ acts as an exact per-step zero-mean baseline that cancels target difficulty without biasing the immediate policy gradient.
(ii) \emph{What does forward consistency PMI optimize---and why does MI-Only RL still need $R_{\text{task}}$?} Expanding the expected forward log-ratio recovers $I_{\nu_\theta}(z_t; \tilde{H}_t)$ minus a conditional critic gap, \emph{plus} $+\KL(P_{\nu_\theta}(z_t) \,\|\, q_\phi(z_t \mid \tilde{H}_{\mathrm{null}}))$; this positive KL term explains why unanchored \textbf{MI-Only RL} ($R_{\text{task}} = 0$, Section~\ref{subsec:ablation}) can inflate consistency PMI by drifting toward telegraphic strings that are improbable under the masked prior $q_\phi(\cdot \mid \tilde{H}_{\mathrm{null}})$.
(iii) \emph{How is offline critic shift controlled?} KL regularization keeps $\nu_\theta$ close to $\nu_{\mathcal{D}}$ early in RL while $\lambda_s = 1$ resolves sparse credit assignment, and cosine annealing ($\lambda_s \to 0$) removes all residual critic and prior-mismatch bias in late training.

\begin{proposition}[PMI Unbiasedness, Recursive Consistency, and Critic Shift]
\label{prop:pmi_nonstationarity}
Let $q_\phi$ be a frozen critic trained on offline distribution $\nu_{\mathcal{D}}$ via Eq.~\eqref{eq:critic_loss}, and let $\nu_\theta$ denote the marginal state-summary distribution under policy $\pi_\theta$. Then:
\begin{enumerate}[label=(\roman*), leftmargin=1.8em, itemsep=2pt]
    \item \textbf{Unbiased Immediate Sufficiency Baseline:} Subtracting the length-normalized masked-summary baseline $\frac{1}{|Y_t|}\log q_\phi(Y_t \mid z_{\mathrm{null}})$ in Eq.~\eqref{eq:r_pred} satisfies $\E_{z_t \sim \pi_\theta(\cdot \mid \tilde{H}_t)}\big[\nabla_\theta \log \pi_\theta(z_t \mid \tilde{H}_t)\,\frac{1}{|Y_t|}\log q_\phi(Y_t \mid z_{\mathrm{null}})\big] = 0$ conditional on $(H_t, Y_t)$ (and across the full trajectory when targets are exogenous), preserving exact per-step policy-gradient unbiasedness for raw $r_{\text{pred}, t}$ while canceling instance-level target entropy.
    \item \textbf{Recursive Consistency Decomposition:} Prior to $1/|z_t|$ length normalization (which rescales per-token values to avoid favoring verbose summaries), the expected sequence-level forward log-ratio satisfies
    \begin{equation}
    \label{eq:forward_pmi_decomp}
    \begin{aligned}
    \E_{\nu_\theta}\!\Big[\log \frac{q_\phi(z_t \mid \tilde{H}_t)}{q_\phi(z_t \mid \tilde{H}_{\mathrm{null}})}\Big] ={} & I_{\nu_\theta}(z_t; \tilde{H}_t) - \E_{\tilde{H}_t}\!\big[\KL(P_{\nu_\theta}(z_t \mid \tilde{H}_t)\,\|\,q_\phi(z_t \mid \tilde{H}_t))\big] \\
    & + \KL\big(P_{\nu_\theta}(z_t)\,\|\,q_\phi(z_t \mid \tilde{H}_{\mathrm{null}})\big).
    \end{aligned}
    \end{equation}
    \item \textbf{KL-Controlled Shift and Annealing:} Relative to reference trajectory distribution $\mu_{\mathrm{ref}}$, policy-to-critic shift satisfies $\TV(\nu_\theta, \nu_{\mathcal{D}}) \le \sqrt{\frac{1}{2}\KL(\mu_\theta \,\|\, \mu_{\mathrm{ref}})} + \TV(\nu_{\mathrm{ref}}, \nu_{\mathcal{D}})$, and decaying $\lambda_s \to 0$ (Eq.~\ref{eq:cosine_anneal}) eliminates residual critic and prior-mismatch bias for $s \ge s_{\mathrm{end}}$.
\end{enumerate}
\end{proposition}
\begin{proof}
\emph{Part (i) and Practical Departures:} Conditional on $(H_t, Y_t)$, neither $|Y_t|$ nor $\log q_\phi(Y_t \mid z_{\mathrm{null}})$ depends on the sampled summary $z_t$, so $\E_{z_t \sim \pi_\theta(\cdot \mid \tilde{H}_t)}\big[\nabla_\theta \log \pi_\theta(z_t \mid \tilde{H}_t)\,\frac{1}{|Y_t|}\log q_\phi(Y_t \mid z_{\mathrm{null}})\big] = \frac{1}{|Y_t|}\log q_\phi(Y_t \mid z_{\mathrm{null}})\,\nabla_\theta \sum_{z_t} \pi_\theta(z_t \mid \tilde{H}_t) = 0$, preserving unbiased immediate policy gradients while subtracting instance-level target difficulty to reduce gradient variance across prompts. In practice, Eq.~\eqref{eq:shaped_step} and Eq.~\eqref{eq:return_to_go} introduce two deliberate departures from a pure linear control variate: (1) in closed-loop dialogue, future baselines $\log q_\phi(Y_k \mid z_{\mathrm{null}})$ ($k > t$) depend on $z_t$ via $a_t$ and normalize future state difficulty so the policy is not rewarded for steering toward trivially low-entropy user states; and (2) passing $r_{\text{pred}, t}$ and $r_{\text{cons}, t}$ through centered sigmoids $\sigma(\alpha(r - b)) - \tfrac{1}{2}$ makes the baseline nonlinear, trading exact variance-reduction linearity for bounded step-wise scale (preventing heavy-tailed log-ratio spikes) and zero-centered length neutrality on variable-horizon episodes. Decaying $\lambda_s \to 0$ (Part~iii) removes any asymptotic shaping bias.

\emph{Part (ii):} Adding and subtracting $\log P_{\nu_\theta}(z_t \mid \tilde{H}_t)$ and $\log P_{\nu_\theta}(z_t)$ inside $\E_{\nu_\theta}[\log q_\phi(z_t \mid \tilde{H}_t) - \log q_\phi(z_t \mid \tilde{H}_{\mathrm{null}})]$ yields Eq.~\eqref{eq:forward_pmi_decomp}. Thus, forward PMI penalizes ungrounded tokens where $q_\phi(z_{t,i} \mid z_{t,<i}, \tilde{H}_t) \approx q_\phi(z_{t,i} \mid z_{t,<i}, \tilde{H}_{\mathrm{null}})$, while the $+\KL(P_{\nu_\theta}(z_t)\,\|\,q_\phi(z_t \mid \tilde{H}_{\mathrm{null}}))$ term explains why purely unanchored MI optimization (\textbf{MI-Only RL}, $R_{\text{task}}=0$) eventually drifts into telegraphic strings unlikely under $q_\phi(\cdot \mid \tilde{H}_{\mathrm{null}})$.

\emph{Part (iii):} By the triangle inequality, data processing for TV (since $\nu_\theta$ is a marginal projection of $\mu_\theta$), and Pinsker's inequality, $\TV(\nu_\theta, \nu_{\mathcal{D}}) \le \TV(\nu_\theta, \nu_{\mathrm{ref}}) + \TV(\nu_{\mathrm{ref}}, \nu_{\mathcal{D}}) \le \sqrt{\frac{1}{2}\KL(\mu_\theta \,\|\, \mu_{\mathrm{ref}})} + \TV(\nu_{\mathrm{ref}}, \nu_{\mathcal{D}})$ (where training $q_\phi$ across high-quality, vague, and hallucinated summary tiers widens $\nu_{\mathcal{D}}$ to reduce $\TV(\nu_{\mathrm{ref}}, \nu_{\mathcal{D}})$). Finally, under Eq.~\eqref{eq:cosine_anneal}, $\lambda_s = 0$ for all $s \ge s_{\mathrm{end}}$, identically removing all residual $\Delta_{\mathrm{post}}(\phi; \theta)$ and masked-prior mismatch terms from Eq.~\eqref{eq:return_to_go} so late-stage PPO updates optimize the exact KL-regularized downstream task objective.
\end{proof}

\section{Synthetic Data Generation}
\label{app:data_generation}

A key advantage of MaxIM is that its dense reward signals are derived entirely from synthetic offline data, eliminating the need for expensive human preference labels. This appendix details the data generation pipelines for both benchmarks.

\subsection{arXiv Data Generation}
\label{app:arxiv_data}

\paragraph{Paper Scraping, Chunk Segmentation, and Disjoint Splits.}
We scrape recent papers from ar5iv~\cite{ar5iv}, the HTML5 mirror of arXiv, in March 2026 and April--September 2026 across 9 diverse categories (\texttt{cs.LG}, \texttt{cs.CL}, \texttt{cs.AI}, \texttt{cs.RO}, \texttt{eess.SY}, \texttt{math.OC}, \texttt{astro-ph.IM}, \texttt{q-bio.QM}, \texttt{physics.comp-ph}) to ensure topical diversity and minimize parametric pretraining leakage. For each paper, we extract the main body prose (excluding the abstract, bibliography, and appendices) and segment it along paragraph and sentence boundaries into $\approx 500$-token chunks under the Gemini SentencePiece tokenizer, yielding up to $25$ chunks per paper. In total, we curate $2{,}196$ papers ($\ge 20$ chunks): $1{,}814$ March 2026 papers partitioned strictly disjointly by salted MD5 hash ($70\% / 30\%$) into $1{,}281$ critic-training papers (used exclusively for offline critic pre-training and Policy SFT) and $533$ RL-training papers (used exclusively for online RL rollouts and Best-of-$N$ MI SFT with $N=8$), $44$ validation papers for checkpoint selection, and $338$ post-cutoff held-out test papers (mean $24.58$ chunks/paper; $5$ synthesis MCQs per paper). For the Early-Section Stress Test, we construct an analogous disjoint split ($525$ RL train / $36$ validation / $332$ test papers) whose comprehension questions are derived strictly from body chunks $3$--$7$ (excluding the Abstract and Introduction).

\paragraph{MCQ Generation and Anti-Leakage Calibration.}
For each paper, we prompt an oracle model (Gemini 3.0 Flash) with the full paper text and request 5 synthesis-level multiple-choice questions requiring cross-section reasoning. Each question has 4 options (A--D) with exactly one correct answer. The prompt explicitly instructs the model to generate questions that cannot be answered from any single section alone, forcing the summarizer to retain cross-sectional information. To prevent frozen downstream solvers from exploiting option-length heuristics or parametric priors without reading the summary, we apply an automated distractor-rewrite pass enforcing $\pm 10\%$ word/character parity across options A--D (with $1$--$2$ distractors slightly longer than the gold answer), deterministically permute option order, and append a fifth abstention option E (\emph{``The summary does not contain enough information to answer this question''}). This calibration reduces no-summary forced-choice accuracy from $67.5\%$ to $27.6\%$ (near the $25\%$ chance floor) and no-summary Abstain-MCQ accuracy to $2.6\%$ ($91.9\%$ abstention). The full MCQ generation prompt is provided in Appendix~\ref{app:prompts}.

\paragraph{Three-Tier Summary Variants and Critic $(\text{Input}, \text{Target})$ Pairs.}
At each turn $t$ of each critic-training paper, we synthesize three summary quality variants using an oracle model:
\begin{enumerate}[leftmargin=1.6em, itemsep=2pt]
    \item \textbf{High-Quality ($z_t^{\text{hq}}$):} Generated with the paper prefix $o_{1:t}$ and MCQ context visible, instructing the oracle to retain the maximal task-relevant evidence within budget $K$. These serve as positive targets for both sufficiency and recursive consistency, as well as Policy SFT supervision.
    \item \textbf{Vague ($z_t^{\text{vague}}$):} Generated by degrading $z_t^{\text{hq}}$ into a fluent but generic overview that omits specific technical mechanisms, equations, and quantitative results, calibrating the critic to penalize information-poor summaries.
    \item \textbf{Hallucinated ($z_t^{\text{halluc}}$):} Generated by altering core claims and empirical findings in $z_t^{\text{hq}}$ into plausible but fabricated statements of matched style and length, calibrating the critic against confabulation.
\end{enumerate}
To construct supervised $(\text{input}, \text{target})$ examples for Eq.~\eqref{eq:critic_loss} (training details in Appendix~\ref{subsec:critic_training}; exact prompt templates in Appendix~\ref{app:critic_prompts}), for every summary $z \in \{z_t^{\text{hq}}, z_t^{\text{vague}}, z_t^{\text{halluc}}, z_{\mathrm{null}}\}$ (where $z_{\mathrm{null}} = \texttt{"(no summary available)"}$), the frozen Abstain-MCQ action model answers the paper's five questions $Q_{1:5}$ from $z$ alone, producing a distilled 5-letter answer sequence $\hat{Y}(z) \in \{\text{A},\dots,\text{E}\}^5$ (selecting E whenever $z$ lacks supporting evidence). Across the $1{,}281$ critic-training papers, $\mathcal{D}$ comprises $136{,}003$ examples across six subsets:
(i) conditional sufficiency pairs $\big((z_t, Q_{1:5}),\, \hat{Y}(z_t)\big)$ for $z_t \in \{z_t^{\text{hq}}, z_t^{\text{vague}}, z_t^{\text{halluc}}\}$ ($26{,}949$, $26{,}949$, and $26{,}926$ examples) plus masked-summary baselines $\big((z_{\mathrm{null}}, Q_{1:5}),\, \hat{Y}(z_{\mathrm{null}})\big)$ ($1{,}281$ examples, one per paper); and
(ii) forward recursive consistency pairs $\big((z_{t-1}^{\text{hq}}, o_t),\, z_t^{\text{hq}}\big)$ ($26{,}949$ examples, with $z_0^{\text{hq}} = \varnothing$) plus masked-context baselines $\big(\tilde{H}_{\mathrm{null}},\, z_t^{\text{hq}}\big)$ ($26{,}949$ examples, replacing both context fields with \texttt{"(not available)"}).

\subsection{WildChat Data Generation}
\label{app:wildchat_data}

\paragraph{Seed Data, Disjoint Partitioning, and Trajectory Regeneration.}
We use WildChat~\cite{zhao2024wildchat} conversations as seeds. Specifically, we extract English conversations with $\ge 6$ user turns ($7{,}968$ validated dialogues) and prompt an annotator model (Gemini 3.0 Flash) sequentially at each turn $t$ with the observable dialogue prefix up to turn $t$ and prior state $s_{t-1}$ to causally annotate the user's evolving internal state. To test out-of-distribution generalization to unseen user personas, $500$ full dialogues are held out from user-simulator training, from which $200$ form our unseen-simulator \textbf{OOD evaluation split}. The remaining $7{,}468$ simulator-training dialogues train the user simulator and are partitioned strictly disjointly into $3{,}000$ critic/SFT seed dialogues ($2{,}928$ rolled out into $25{,}497$ turns for critic and Policy SFT training, plus $2{,}511$ held-out critic validation triplets), $3{,}500$ RL seed dialogues for online RL, $200$ validation dialogues, $500$ held-out test dialogues, and $268$ offline PMI-calibration seeds. Because the original WildChat conversations were generated with an assistant model to which we do not have access, we synthetically regenerate full multi-turn trajectories using an action agent and a trained user simulator (Appendix~\ref{subsec:user_sim}). This creates a controlled closed-loop environment where we have access to ground-truth user internal states at every turn.

\paragraph{Dynamic Conversation Generation.}
Each conversation proceeds for up to $12$ turns during RL training and up to $15$ turns at evaluation (testing $+25\%$ horizon extrapolation), initialized with $z_0 = \varnothing$ and persona $s_0$, via the following loop for turn $t = 1, 2, \dots$:
\begin{enumerate}[leftmargin=1.6em, itemsep=2pt]
    \item The user simulator generates the next user utterance $u_t$ conditioned on prior latent state $s_{t-1}$ and conversation history. If the simulator outputs a termination signal, the episode ends.
    \item The action agent generates a response $a_t$ conditioned on the conversation context and the summarizer's prior belief estimate $z_{t-1}$ injected into its system prompt (no belief injected at $t=1$, since $z_0 = \varnothing$).
    \item Given user observation $o_t = u_t$ and assistant action $a_t$, the summarizer policy updates the post-turn belief $z_t \sim \pi_\theta(\cdot \mid z_{t-1}, o_t, a_t)$ (post-action turn indexing so $z_t$ aligns with post-turn user state $s_t$), and the user simulator updates $s_t$ ($Y_t$).
\end{enumerate}

\paragraph{Belief Summary Variants and 10-Key Schema.}
Analogous to the arXiv pipeline, we generate three quality tiers of belief summaries (HQ, vague, hallucinated) plus masked-context baseline queries ($z_{\mathrm{null}}, \tilde{H}_{\mathrm{null}}$) at each turn using an oracle model with access to the true user internal states. Belief states are structured as canonical 10-key probabilistic JSON objects inside \texttt{<state>...</state>} tags: \texttt{core\_\allowbreak intent\_\allowbreak functional\_\allowbreak goal}, \texttt{affective\_\allowbreak state\_\allowbreak perceived\_\allowbreak stakes}, \texttt{user\_\allowbreak expertise\_\allowbreak epistemic\_\allowbreak stance}, \texttt{cognitive\_\allowbreak demand\_\allowbreak information\_\allowbreak specificity}, \texttt{interaction\_\allowbreak strategy\_\allowbreak communication\_\allowbreak demeanor}, \texttt{agent\_\allowbreak perception\_\allowbreak trust\_\allowbreak output\_\allowbreak control}, \texttt{task\_\allowbreak state\_\allowbreak domain\_\allowbreak context}, \texttt{patience\_\allowbreak reserve\_\allowbreak accumulator}, \texttt{task\_\allowbreak success\_\allowbreak outlook}, and \texttt{specific\_\allowbreak needs\_\allowbreak and\_\allowbreak constraints}. For vague sufficiency examples, four keys are degraded to generic placeholders paired with \texttt{<insufficient\_information>} targets; for hallucinated sufficiency examples, four keys receive orthogonal distractor traits, enforcing strict critic conditioning on $z_t$.

\section{MaxIM Training Algorithm}
\label{app:algorithm}

Algorithm~\ref{alg:maxim} details the complete three-phase pipeline for training the MaxIM agent, including offline synthetic data generation, unified critic pre-training, and KL-regularized PPO policy optimization with mid-training cosine annealing.

\begin{algorithm}[htbp]
\caption{MaxIM: Maximally Informative Incremental Memory Update}
\label{alg:maxim}
\begin{algorithmic}[1]
\REQUIRE Initial Policy $\pi_\theta$, unified critic $q_\phi$, Weights $(w_s, w_c)$, Discount $\gamma$, Shaping scale $\eta_{\text{shape}}$, Annealing schedule $\lambda_s$, Token limit $K$
\STATE \textbf{Phase 1: Synthetic Data Generation}
\STATE Generate offline trajectories $\mathcal{D}$ with privileged access to target $Y$ on the disjoint critic-training split.
\STATE \textbf{Phase 2: Critic Pre-training \& Offline PMI Calibration}
\STATE Train $q_\phi$ via supervised cross-entropy minimization (Eq.~\ref{eq:critic_loss}) on $\mathcal{D}$ and calibrate $(\alpha_s, b_s, \alpha_c, b_c)$ on initial base-policy rollouts.
\STATE \textbf{Phase 3: RL Fine-Tuning}
\WHILE{training step $s = 1 \dots S_{\max}$}
    \STATE Initialize trajectory buffer $\mathcal{B} \leftarrow \emptyset$
    \FOR{each initial environment state in batch ($8$ prompts)}
        \STATE Sample a group of $G=8$ trajectories from the same initial conditions:
        \FOR{each trajectory $i \in \{1, \dots, G\}$ in group}
            \STATE $z_0 \leftarrow \text{Empty String}$, $a_0 \leftarrow \text{None}$
            \FOR{$t = 1 \dots T_i$}
                \STATE Observe $o_t$ (in interactive dialogue, assistant first replies $a_t \sim \pi_{\text{act}}(\cdot \mid z_{t-1}, o_t, \dots)$) and record privileged target $Y_t$.
                \STATE Sample $z_t \sim \pi_\theta(\cdot \mid \tilde{H}_t)$ subject to $|z_t| \le K$, where $\tilde{H}_t = (z_{t-1}, a_{t-1}, o_t)$ (or post-turn $\tilde{H}_t = (z_{t-1}, o_t, a_t)$ in dialogue; $(z_{t-1}, o_t)$ in passive reading).
                \STATE Store state-action tuple $(\tilde{H}_t, z_t)$.
            \ENDFOR
            \STATE Evaluate terminal action (if deferred to $T_i$, e.g., MCQ answering from $z_{T_i}$) and receive terminal task reward $R_{\text{task}, T_i}$ ($R_{\text{task}, t} = 0$ for $t < T_i$).
            \STATE Initialize $G_{T_i + 1}^{(i)} = 0$.
            \FOR{$t = T_i \dots 1$}
                \STATE Compute PMI scores $r_{\text{pred}, t}, r_{\text{cons}, t}$ (Eqs.~\ref{eq:r_pred}--\ref{eq:r_cons}) via teacher-forced critic passes.
                \STATE Compute centered sigmoid MI step reward (Eq.~\ref{eq:shaped_step}): $\hat{r}_t = w_s\big(\sigma(\alpha_s(r_{\text{pred},t}-b_s))-\tfrac{1}{2}\big) + w_c\big(\sigma(\alpha_c(r_{\text{cons},t}-b_c))-\tfrac{1}{2}\big)$.
                \STATE Compute discounted return: $G_t^{(i)} = R_{\text{task}, t} + \eta_{\text{shape}} \lambda_s\, \hat{r}_t + \gamma G_{t+1}^{(i)}$.
                \STATE Append transition $(\tilde{H}_t, z_t, G_t^{(i)})$ to group buffer.
            \ENDFOR
        \ENDFOR
        \STATE Standardize returns $G_t^{(i)}$ via turn-indexed leave-one-out baseline and prompt-wide std scaling to compute advantages $A_t^{(i)}$, append to $\mathcal{B}$.
    \ENDFOR
    \STATE Update policy $\pi_\theta$ using KL-regularized PPO with advantage estimates in $\mathcal{B}$.
\ENDWHILE
\end{algorithmic}
\end{algorithm}

\section{Details of Experiments}
\label{app:experiments}

\subsection{User Simulator}
\label{subsec:user_sim}

The WildChat user simulator is trained via SFT on the offline WildChat dataset~\cite{zhao2024wildchat}, where per-turn internal state annotations were first generated by an oracle annotator model. Specifically, we train a unified \textbf{joint state-and-response user simulator} (initialized from Gemini 2.5 Flash and fine-tuned for $6$ epochs on the $7{,}468$ simulator-training dialogues). At each turn, conditioned on the initial persona $s_0$ and full dialogue history, the simulator emits a structured trace \texttt{<think><impact>...</impact><dimensions>\{10-key JSON\}</dimensions></think>} followed immediately by the next natural-language user utterance (or a termination token) in a single call. Up to two schema-validation resamples ensure $100\%$ canonical 10-key ground-truth state coverage across multi-turn rollouts (median length $8.0$ turns).

\subsection{Critic Training}
\label{subsec:critic_training}

The variational critic $q_\phi$ is trained via standard cross-entropy minimization over target tokens only (masking prompt and instruction tokens). Rather than maintaining separate checkpoints for sufficiency and consistency, each benchmark trains a \textbf{single unified Gemini 2.5 Flash critic} on a six-way multi-task mixture (assigning unit sampling weight $1.0$ to HQ, vague, and hallucinated conditional sufficiency examples as well as conditional and masked-context grounding examples, and weight $0.05$ to masked-summary sufficiency baselines):
\begin{itemize}[leftmargin=1.4em, itemsep=2pt]
    \item \textbf{Sufficiency PMI ($r_{\text{pred}, t}$):} On arXiv, $q_\phi$ scores the mean log-likelihood of the $5$ single-token Abstain-MCQ target letters (A--E) conditioned on $z_t$ minus the cached masked-summary baseline ($z_t = z_{\mathrm{null}}$). On WildChat, $q_\phi$ scores the per-token log-likelihood of the canonical 10-key post-turn true user state $s_t$ conditioned on $z_t$ minus the masked-state baseline ($z_{\mathrm{null}}$).
    \item \textbf{Recursive Consistency PMI ($r_{\text{cons}, t}$):} On both benchmarks, $q_\phi$ scores the per-token log-likelihood of the generated summary $z_t$ conditioned on local context $\tilde{H}_t$ ($\tilde{H}_t = (z_{t-1}, o_t)$ on arXiv; $\tilde{H}_t = (z_{t-1}, o_t, a_t)$ on WildChat) minus its masked-context baseline conditioned on $\tilde{H}_{\mathrm{null}}$.
\end{itemize}
Training uses three quality tiers---HQ, vague, and hallucinated---so the critic sees the full spectrum of summary quality during training. We use single-epoch offline training ($\le 1$ epoch over HQ targets) to prevent memorization. The same critic checkpoint is shared across all shaped RL conditions within a benchmark. While we train one unified critic per benchmark here for clean experimental isolation, because Eq.~\eqref{eq:critic_loss} routes tasks purely through natural-language system instructions over token cross-entropy, the same architecture directly supports joint multi-task and cross-domain critic training.

\subsection{RL Hyperparameters and Implementation Details}
\label{app:implementation-details}

\paragraph{Reward Transformation, Offline Calibration, and Annealing.}
To combine sufficiency and consistency PMI scores on a common bounded scale while centering neutral summaries at zero, we evaluate initial pre-RL base-policy rollouts strictly offline to compute the mean midpoint $b$ and within-prompt standard deviation $\sigma_{\text{within}}$ for each PMI signal, setting slope $\alpha = 1/\sigma_{\text{within}}$ and mapping each signal to $\sigma(\alpha(r - b)) - \tfrac{1}{2} \in [-0.5, +0.5]$ (eliminating positive survival-length bias on variable-horizon WildChat dialogues). On arXiv ($K=500$), this yields $(b_s, b_c) = (-0.28, 1.71)$ and $(\alpha_s, \alpha_c) = (2.85, 7.95)$ (with budget-specific midpoints $b_s \in \{-0.79, -0.43, -0.28, -0.10\}$ and $b_c \in \{2.02, 1.78, 1.71, 1.68\}$ for $K \in \{100, 300, 500, 800\}$, and $b_s = -0.72$ on the Early-Section Stress Test); on WildChat, $(b_s, b_c) = (0.0921, 0.4950)$ and $(\alpha_s, \alpha_c) = (33.53, 4.36)$. Across both domains, we use MI weights $(w_s, w_c) = (0.30, 0.11)$ alongside $R_{\text{task}} \in [0, 1]$, discount $\gamma = 0.9$, step-shaping scale $\eta_{\text{shape}} = 1 - \gamma = 0.1$ to account for multi-turn reward accumulation, and cosine annealing $\lambda_s$ holding $1.0$ through step $s_{\text{start}} = 450$ before decaying smoothly to $0.0$ at step $s_{\text{end}} = 600$:
\begin{equation}
\label{eq:cosine_anneal}
\lambda_s = \begin{cases}
1, & s \le s_{\text{start}}, \\
\frac{1}{2}\!\left(1 + \cos\!\left(\pi \frac{s - s_{\text{start}}}{s_{\text{end}} - s_{\text{start}}}\right)\right), & s_{\text{start}} < s < s_{\text{end}}, \\
0, & s \ge s_{\text{end}}.
\end{cases}
\end{equation}
In addition, to enforce valid XML delimiters during RL across all methods, a lightweight tag-formatting penalty (weight $0.05$, scaled by $0.1$) assigns $0$ when complete \texttt{<summary>...</summary>} (arXiv) or \texttt{<state>...</state>} (WildChat) tags are present, $-0.5$ if only one tag is present, and $-1.0$ if neither is present.

\paragraph{Advantage Normalization.}
Table~\ref{tab:hyperparams} summarizes our training and evaluation settings. For group-relative advantage estimation across $G=8$ rollouts per prompt, let $\mathcal{G}_t = \{j \in \{1,\dots,G\} : T_j \ge t\}$ denote active trajectories at turn $t$. Whenever $|\mathcal{G}_t| \ge 2$, we compute an exact leave-one-out baseline $\tilde{A}_{i,t} = G_t^{(i)} - \frac{1}{|\mathcal{G}_t|-1}\sum_{j \in \mathcal{G}_t \setminus \{i\}} G_t^{(j)}$ (setting $\tilde{A}_{i,t}=0$ if $|\mathcal{G}_t| < 2$) and normalize by the prompt-wide standard deviation across all valid turn-rollout pairs in the group, avoiding noisy small-sample variance division at late turns. Policies are updated using KL-regularized PPO~\cite{schulman2017proximal}.

\begin{table}[htbp]
\centering
\caption{\textbf{Summary of training and evaluation settings.} Settings shared identically across both benchmarks span a single combined column.}
\label{tab:hyperparams}
\vspace{2pt}
{\footnotesize
\setlength{\tabcolsep}{5.0pt}
\begin{tabular}{lcc}
\toprule
\textbf{Setting / Hyperparameter} & \textbf{arXiv Scientific Reading} & \textbf{WildChat Multi-Turn Dialogue} \\
\midrule
Base Backbone (Policy, Critic, Actor) & \multicolumn{2}{c}{Gemini 2.5 Flash} \\
Offline Target / RL Belief Rater & \multicolumn{2}{c}{Gemini 3.0 Flash} \\
Atomic-Claim Splitter & \multicolumn{2}{c}{Gemini 2.5 Flash-Lite} \\
Prompts per Step $\times$ Group Size $G$ & \multicolumn{2}{c}{$8 \times 8 = 64$ trajectories/step} \\
Offline SFT Batch Size (Critic / Policy SFT) & \multicolumn{2}{c}{$32$--$64$ ($\le 1$ epoch over HQ targets)} \\
Discount $\gamma$ / Step Scale $\eta_{\text{shape}}$ & \multicolumn{2}{c}{$0.9$ / $0.1$ (MemAgent RL: $\gamma{=}1.0$)} \\
MI Shaping Weights $(w_s, w_c)$ & \multicolumn{2}{c}{$(0.30, 0.11)$} \\
MI Cosine-Decay Window $(s_{\text{start}} \to s_{\text{end}})$ & \multicolumn{2}{c}{RL steps $450 \to 600$} \\
\midrule
External Grounding \& Qualitative Judge & GPT-4.1 (+ Gemini 3.0 Flash check) & GPT-4.1 ($k{=}1$ eval window) \\
Summary Token Budget $K$ (Train / Eval) & $500$ (sweep: $100, 300, 500, 800$) & $1{,}024$ \\
Episode Horizon $T$ (Train / Eval) & $20$ train / $25$ eval chunks & $12$ train / $15$ eval turns \\
Total Online RL Steps $S_{\max}$ & $1{,}000$ ($45$--$60\%$ decay; ckpt $/25$) & $800$ ($56$--$75\%$ decay; ckpt $/25$) \\
Calibrated Midpoints $(b_s, b_c)$ & $(-0.28, 1.71)$ at $K{=}500$ & $(0.0921, 0.4950)$ \\
Calibrated Slopes $(\alpha_s, \alpha_c)$ & $(2.85, 7.95)$ & $(33.53, 4.36)$ \\
\bottomrule
\end{tabular}%
}
\end{table}

\paragraph{Optimization and Computational Overhead.}
Computing the step-wise variational rewards requires only teacher-forced forward passes through the critic transformer, avoiding expensive autoregressive generation. To ensure we do not block environment trajectory generation, critic reward computation is executed asynchronously across distributed inference servers and collected at episode end.

\subsection{Evaluation Protocol, Training Compute, and Inference Cost}
\label{app:compute_cost}

\paragraph{Task Metrics, Bounded-Memory Multi-Turn Dialogue, and Causal Belief Scoring.}
For arXiv, a frozen action model answers 5 MCQs conditioned only on the final summary. Under our primary \textbf{Abstain-MCQ} protocol (options A--E), correct answers earn $+1$, abstentions (E) earn $0$, and wrong commitments (A--D) incur $-1/3$. For WildChat, because evaluating a conversational agent spans multiple distinct quality axes (goal fulfillment, constraint adherence, coherence) and collapsing them into a single scalar reward risks misspecification, during RL training an oracle rater (Gemini 3.0 Flash) receives $(s_0, u_1, a_1, s_1, z_1, \dots, s_T, z_T)$ under a strict \textbf{causal $1\dots t$ grounding rule} (never penalizing $z_t$ for traits revealed at turns $> t$) and outputs $1$--$10$ belief accuracy ratings (rescaled to $[0,1]$ as $R_{\text{task}}$, and reported in Table~\ref{tab:wc_belief_fidelity}). At test time (RQ1, Tables~\ref{tab:main_rq1} and~\ref{tab:wc_downstream_full}), to evaluate downstream conversational utility under bounded context without retraining the assistant, we execute fresh rollouts ($T \le 15$) on the $N=500$ held-out test split where the frozen Gemini 2.5 Flash assistant receives only the single preceding user--assistant interaction ($k=1$: $(u_{t-1}, a_{t-1})$) and the latest user message $u_t$ together with $z_{t-1}$ (whereas the \textbf{No-Belief} baseline receives only $(u_{t-1}, a_{t-1}, u_t)$). An independent \textbf{GPT-4.1} evaluator ($\text{temperature}=0$), conditioned strictly on $(u_{1:T}, a_{1:T})$ and ground-truth user states $s_{0:T}$ without access to $z_t$ or method identity, rates each session ($1$--$10$) on \textbf{Goal Completion}, \textbf{Preference \& Constraint Retention}, \textbf{Coherence \& Adaptability}, and \textbf{Overall User Satisfaction}. Statistical significance between paired conversation seeds is computed via $10{,}000$-resample paired bootstrap on both all $N=500$ seeds and the fixed $T \ge 10$ cohort ($N=158$ seeds where No-Belief requires $T \ge 10$ turns).

\paragraph{Qualitative Autorater and Atomic-Claim Grounding Protocols.}
For qualitative evaluation (Table~\ref{tab:grounding_rq2}, Appendix~\ref{app:turnwise_e4}), a cross-family \textbf{GPT-4.1} judge ($\text{temperature}=0$) compares candidate summaries conditioned on identical observed prefixes up to turn $t$ with randomized A/B order on \textbf{Truthfulness} and \textbf{Usefulness}, and independently scores each policy's single-step transition $(z_{t-1}, o_t) \to z_t$ on a $1$--$10$ \textbf{Consistency} scale so ratings are not confounded by differences in $z_{t-1}$ (prompts in Appendix~\ref{app:prompts}). While our evaluation pipeline computes all three qualitative metrics on both tasks, Table~\ref{tab:grounding_rq2} reports Consistency on arXiv and Truthfulness on WildChat due to complementary ceiling effects: on arXiv prose summaries, factual precision saturates near ceiling across RL methods ($97.5\%$--$98.4\%$ entailment) whereas Consistency strongly separates policies ($5.99$--$7.55$); conversely, on WildChat's fixed 10-key JSON schema, standalone Consistency saturates near ceiling across all RL policies ($9.69$--$9.86$; Appendix~\ref{app:turnwise_e4}) whereas Truthfulness varies substantially ($36.2\%$--$72.4\%$). For atomic-claim grounding, \textbf{Gemini 2.5 Flash-Lite} ($\text{temperature}=0$) first decomposes each candidate summary $z_t$---and, for \emph{chunk claim incorporation} (Figure~\ref{fig:scaling}(b), Table~\ref{tab:fig2b_incorp}), each incoming source chunk $o_t$ (extracted once per paper and shared across all methods)---into standalone, de-contextualized atomic factual claims; GPT-4.1 then classifies each claim against the observed source prefix as \texttt{ENTAILED}, \texttt{NEUTRAL}, or \texttt{CONTRADICTED} via natural language inference (NLI), complemented by a Gemini 3.0 Flash cross-check on arXiv (Appendix~\ref{app:turnwise_e4}).

\paragraph{Training and Inference Computational Complexity.}
All SFT, RL, and serving jobs run on Google Cloud TPU accelerators. Let $S_{\max}$ denote the total number of online RL update steps ($S_{\max} = 1{,}000$ on arXiv and $S_{\max} = 800$ on WildChat) and $N_{\text{param}}$ denote the active parameter count of the shared Gemini 2.5 Flash backbone used across every trainable policy, unified critic, action agent, and user simulator. Because shared-cluster wall-clock duration fluctuates with serving-replica concurrency, Table~\ref{tab:compute_cost} reports hardware-independent \textbf{relative training FLOPs} normalized to sparse-reward RL ($1.00\times$) under standard transformer accounting ($2N_{\text{param}}$ FLOPs per forward/decode token, $6N_{\text{param}}$ per forward-backward training token, plus $2N_{\text{param}}$ PPO reference evaluation). Because critic passes are teacher-forced log-likelihood evaluations ($2N_{\text{param}}$/token, zero decoding), masked baselines are cached once per episode, and cosine annealing sets $\lambda_s = 0$ for $s \ge 600$, MaxIM requires only $\approx 1.25\times$ total training FLOPs on arXiv ($1.23\times$ online RL plus $0.02\times$ one-time critic SFT) and $\approx 1.17\times$ on WildChat ($1.14\times$ online RL plus $0.03\times$ critic SFT) relative to Task-Only/MemAgent RL, adding $< 15\%$ per-step RL wall-clock overhead during $s < 600$ ($0\%$ for $s \ge 600$). Beyond the final benchmark runs in Table~\ref{tab:compute_cost}, preliminary development (data-pipeline validation, reward-weight and annealing sweeps) consumed roughly $2.5\times$ the compute of the reported experiments. At deployment, \textbf{test-time inference incurs zero critic overhead ($1.00\times$ FLOPs and latency)}: every method executes the same single summarizer call per turn ($O(T \cdot K)$ generated tokens), avoiding the $O(T^2 L_{\text{chunk}})$ quadratic attention cost of full-context prompting.

\begin{table}[htbp]
\centering
\caption{\textbf{Training-time forward/generation budget, relative training FLOPs ($1.00\times =$ sparse-reward RL), and test-time inference cost per $T$-turn episode.} $S_{\max}$ is total RL update steps ($1{,}000$ on arXiv; $800$ on WildChat). Critic passes use teacher-forced log-likelihoods ($0$ generated tokens).}
\label{tab:compute_cost}
\vspace{2pt}
{\footnotesize
\setlength{\tabcolsep}{3.2pt}
\begin{tabular}{lccccc@{\hspace{10pt}}c}
\toprule
\textbf{Method} & \mcell{\textbf{Offline SFT}\\\textbf{Steps}} & \mcell{\textbf{Online RL}\\\textbf{Trajectories}} & \mcell{\textbf{Critic Passes}\\\textbf{per RL Turn}} & \mcell{\textbf{Rel.\ Train}\\\textbf{FLOPs}} & \mcell{\textbf{Test}\\\textbf{Critic}} & \mcell{\textbf{Test Memory}\\\textbf{Calls / FLOPs}} \\
\midrule
Prompt-Only & $0$ & $0$ & $0$ & $0$ & $\mathbf{0}$ & \multirow{10}{*}{\mcell{$T$ calls ($1.00\times$)\\$\le K$ tok/turn\\(identical across\\all methods)}} \\
\addlinespace[3pt]
Policy SFT & \mcell{$1{,}000$ (arXiv)\\$600$ (WC)} & $0$ & $0$ & $\approx 0.02\times$ & $\mathbf{0}$ & \\
\addlinespace[3pt]
\lcell{Best-of-$N$ MI SFT\\($N{=}8$, arXiv)} & \mcell{$350$ (+ $8{\times}533$\\offline rollouts)} & $0$ & \mcell{$2$ cond.\ +\\cached base.} & $\approx 0.08\times$ & $\mathbf{0}$ & \\
\addlinespace[3pt]
\lcell{MemAgent RL /\\Task-Only RL} & $0$ & \mcell{$64 {\times} S_{\max}$\\($64\text{K}$ arXiv /\\$51.2\text{K}$ WC)} & $0$ & $1.00\times$ & $\mathbf{0}$ & \\
\addlinespace[3pt]
\lcell{\textbf{MaxIM}\\\textbf{(Full, Ours)}} & \mcell{$1{,}000$ / $900$\\(Critic SFT)} & \mcell{$64 {\times} S_{\max}$\\($64\text{K}$ arXiv /\\$51.2\text{K}$ WC)} & \mcell{$2$ cond.\ + cached\\base.\ ($s < 600$)} & \mcell{$\approx 1.25\times$ (arXiv)\\$\approx 1.17\times$ (WC)} & $\mathbf{0}$ & \\
\bottomrule
\end{tabular}%
}
\end{table}

\section{Prompt Templates}
\label{app:prompts}

This appendix documents the prompt templates used across MaxIM's summarizer policies, synthetic data pipelines, variational critics, and evaluation judges. Template variables injected at runtime are denoted in monospace braces (e.g., \pvar{max_summary_tokens}).

\subsection{Incremental Summarizer Policy Prompts}

At each turn $t$, the summarizer policy $\pi_\theta$ receives a domain-specific system prompt and a user message containing the previous summary $z_{t-1}$ alongside the latest chunk or dialogue exchange. On WildChat, the output inside \texttt{<state>...</state>} follows the 10-key JSON schema defined in Appendix~\ref{app:wildchat_data}.

\begin{promptbox}{arXiv Incremental Summarizer Prompt}
\textbf{System:} You are reading a scientific paper one chunk at a time. After each chunk of text, you must update your running summary of the paper. Constraint: Your summary must be at most \pvar{max_summary_tokens} tokens. Your goal is to retain the most important information for understanding and reasoning about the paper. Wrap your summary in \texttt{<summary>...</summary>} tags.\\[3pt]
\textbf{User:} Previous summary: \pvar{prev_summary_or_empty}\quad New paper chunk (\pvar{turn}): \pvar{chunk_text}
\end{promptbox}

\begin{promptbox}{WildChat Incremental Belief-State Summarizer Prompt}
\textbf{System:} You are estimating a user's hidden internal state during a conversation between the user and an AI assistant. Maintain a probabilistic belief about the user's internal state across the canonical 10-key JSON schema, expressing uncertainty when appropriate (e.g., ``likely frustrated'', ``possibly confused''). Wrap your JSON state estimate in \texttt{<state>...</state>} tags (at most \pvar{max_state_tokens} tokens).\\[3pt]
\textbf{User:} Previous belief state: \pvar{prev_state_or_empty}\quad Latest exchange (Turn \pvar{turn}): User: \pvar{u_t} / Assistant: \pvar{a_t}
\end{promptbox}

\subsection{Offline Benchmark Synthesis Prompt}

To construct cross-section comprehension targets for arXiv (Appendix~\ref{app:arxiv_data}), Gemini 3.0 Flash generates five synthesis-level MCQs from the full paper text prior to distractor length-parity calibration and option-E augmentation.

\begin{promptbox}{arXiv Cross-Section MCQ Synthesis Prompt (Gemini 3.0 Flash)}
\textbf{System / User:} You are an expert scientific reviewer. Given the full text of an academic paper, generate exactly 5 synthesis-level multiple-choice questions (MCQs) that test deep comprehension of the paper.\\
Requirements: (1) Each question must require synthesizing information from multiple sections and must NOT be answerable from surface-level single-paragraph recall. (2) Each question must have 4 options (A--D) with exactly 1 correct answer and a verification reasoning trace. (3) Cover core problem motivation, methodological mechanisms, empirical findings, limitations, and cross-section connections.\\[2pt]
Paper text: \pvar{full_paper_text}
\end{promptbox}

\subsection{Unified Critic Prompts for PMI Estimation}
\label{app:critic_prompts}

As described in Section~\ref{subsec:objective} and Appendix~\ref{subsec:critic_training}, a single unified critic checkpoint $q_\phi$ per benchmark handles all four conditional and null-baseline log-likelihood queries in Eqs.~\eqref{eq:r_pred}--\eqref{eq:r_cons}. Cross-entropy training loss and RL teacher-forced scoring are applied strictly to the \textbf{Target} tokens; null baselines replace the conditioning summary or local context with fixed placeholder strings (\texttt{"(no summary available)"}, \texttt{"(no state estimate available)"}, or \texttt{"(not available)"}).

\begin{promptbox}{Unified Critic: arXiv Predictive Sufficiency ($q_\phi(Y_t \mid z_t)$ and $q_\phi(Y_t \mid z_{\mathrm{null}})$)}
\textbf{System:} Given a running summary of a scientific paper and 5 multiple-choice questions (options A--D plus E: insufficient information), predict the answer letter selected from the summary alone.\\[2pt]
\textbf{User:} Summary: \pvar{z_t_or_"(no summary available)"}\quad Questions: \pvar{five_mcqs_with_options_A_to_E}\\[2pt]
\textbf{Target (scored tokens):} \texttt{Answer 1:} \pvar{a_1}\texttt{\textbackslash nAnswer 2:} \pvar{a_2}\texttt{\textbackslash nAnswer 3:} \pvar{a_3}\texttt{\textbackslash nAnswer 4:} \pvar{a_4}\texttt{\textbackslash nAnswer 5:} \pvar{a_5}
\end{promptbox}

\begin{promptbox}{Unified Critic: WildChat Predictive Sufficiency ($q_\phi(Y_t \mid z_t)$ and $q_\phi(Y_t \mid z_{\mathrm{null}})$)}
\textbf{System:} Given the estimated user belief state, predict the canonical 10-key ground-truth user internal state JSON.\\[2pt]
\textbf{User:} Estimated state: \pvar{z_t_or_"(no state estimate available)"}\\[2pt]
\textbf{Target (scored tokens):} \texttt{True state:} \pvar{canonical_10key_true_state_json}
\end{promptbox}

\begin{promptbox}{Unified Critic: Forward Recursive Consistency ($q_\phi(z_t \mid \tilde{H}_t)$ and $q_\phi(z_t \mid \tilde{H}_{\mathrm{null}})$)}
\textbf{System:} Given the previous summary/state and the latest chunk or dialogue turn, predict the updated summary/state.\\[2pt]
\textbf{User:} Previous state: \pvar{z_prev_or_"(not available)"}\quad Latest chunk/turn: \pvar{o_t_exchange_or_"(not available)"}\\[2pt]
\textbf{Target (scored tokens):} \texttt{Updated state:} \pvar{z_t}
\end{promptbox}

\subsection{Downstream Task Evaluation Prompts}

At test time, arXiv terminal summaries $z_T$ are evaluated by the frozen Abstain-MCQ action model, and WildChat bounded-memory ($k=1$) conversations are scored by an independent GPT-4.1 judge that never observes $z_t$.

\begin{promptbox}{Frozen Abstain-MCQ Action Model Prompt (arXiv)}
\textbf{System / User:} Answer each multiple-choice question using ONLY the provided paper summary. Choose A, B, C, or D ONLY if the summary contains sufficient supporting evidence; otherwise select E (``The summary does not contain enough information to answer this question''). Scoring awards $+1$ for a correct A--D answer, $0$ for E, and $-1/3$ for an incorrect A--D guess.\\[2pt]
Summary: \pvar{z_T}\quad Questions: \pvar{five_mcqs_with_options_A_to_E}
\end{promptbox}

\begin{promptbox}{Bounded-Memory Downstream Chat Evaluator Prompt (WildChat, GPT-4.1)}
\textbf{System / User:} You are an impartial evaluator assessing an AI assistant in a multi-turn conversation. Given ONLY (1) the complete User/Assistant transcript \pvar{u_1..T, a_1..T} and (2) the user's ground-truth internal states \pvar{s_0..T} (with no access to any model belief state), rate the Assistant on a $1$--$10$ integer scale with turn-level evidence on four axes:
(1) \textbf{Goal Completion} (fulfillment of the user's underlying functional goal);
(2) \textbf{Preference \& Constraint Retention} (adherence to earlier user preferences and constraints);
(3) \textbf{Coherence \& Adaptability} (non-repetitive progression adapted to user feedback); and
(4) \textbf{Overall User Satisfaction} (holistic session-level utility).
\end{promptbox}

\subsection{Atomic-Claim Grounding and Qualitative Autorater Prompts}

Our source-grounding pipeline (Section~\ref{subsec:autorater}, Appendix~\ref{app:compute_cost}) first decomposes passages into standalone atomic claims using Gemini 2.5 Flash-Lite and then verifies source entailment in batched NLI calls using GPT-4.1 (or Gemini 3.0 Flash). Qualitative evaluation uses GPT-4.1 for randomized A/B pairwise judging (Truthfulness, Usefulness) and standalone $1$--$10$ Consistency scoring.

\begin{promptbox}{Atomic Factual Claim Decomposition Prompt (Gemini 2.5 Flash-Lite)}
\textbf{User:} You are an expert natural language processing auditor. Break down the provided passage into a list of standalone, self-contained atomic claims/facts. Guidelines: (1) Each claim must be a complete, grammatical sentence expressing a single fact or assertion. (2) De-contextualize claims so they are self-contained (replace pronouns such as ``they'' or ``it'' with explicit nouns). (3) Do NOT omit any factual information from the passage. Return ONLY a valid JSON list of strings.\\[2pt]
Passage to decompose: \pvar{summary_z_t_or_source_chunk_o_t}
\end{promptbox}

\begin{promptbox}{Atomic-Claim Source Grounding NLI Prompt (GPT-4.1 / Gemini 3.0 Flash)}
\textbf{User:} You are an expert fact-checker and Natural Language Inference (NLI) auditor. For each numbered Hypothesis claim below, determine its relationship to the Premise (observed source chunks/turns $1\dots t$): \texttt{ENTAILED} (fully supported and directly verified by the premise), \texttt{CONTRADICTED} (contradicts or distorts facts in the premise), or \texttt{NEUTRAL} (contains ungrounded claims or assumptions not supported by the premise). Output \texttt{Claim k: <1-sentence explanation> | Verdict: <ENTAILED|CONTRADICTED|NEUTRAL>}.\\[2pt]
Premise: \pvar{observed_source_prefix_1..t}\quad Hypotheses: \pvar{numbered_atomic_claims}
\end{promptbox}

\begin{promptbox}{Pairwise Truthfulness and Usefulness Autorater Prompt (arXiv \& WildChat, GPT-4.1)}
\textbf{User:} Compare candidate summaries/belief states \textbf{A} and \textbf{B} (randomized order) at turn \pvar{t} given the observed history $1\dots \pvar{t}$ and task reference context (paper MCQs on arXiv; true user states $s_{1:\pvar{t}}$ on WildChat). Evaluate two criteria independently:
(1) \textbf{Truthfulness:} Which candidate is more factually grounded in the observed history up to turn \pvar{t}, avoiding unsupported or fabricated claims, intents, or constraints?
(2) \textbf{Usefulness:} Which candidate retains more task-relevant information for downstream question answering (arXiv) or crafting helpful assistant replies (WildChat)?
Output per axis: \texttt{Reasoning: <...> | Preference: <A, B, or Tie>}.
\end{promptbox}

\begin{promptbox}{Standalone $1$--$10$ Consistency Autorater Prompt (GPT-4.1)}
\textbf{User:} Given (1) the previous summary/belief state $z_{t-1}$ after turn \pvar{t-1}, (2) the new observation/exchange at turn \pvar{t}, and (3) the updated summary/belief state $z_t$, rate \textbf{Consistency} on a $1$--$10$ integer scale: Does $z_t$ preserve valid prior information from $z_{t-1}$, faithfully integrate relevant new evidence from turn \pvar{t}, and evolve coherently without abrupt drift? Output: \texttt{Reasoning: <...> | Score: <1..10>}.
\end{promptbox}

\section{Extended Empirical Results}
\label{app:extended_results}
This appendix provides the complete numerical tables and turn-level diagnostics supporting Figure~\ref{fig:scaling} and Sections~\ref{subsec:objective} and~\ref{sec:experiments}.

\subsection{Full Token-Budget Scaling and Chunk Claim Incorporation}
\label{app:e2_table}

Table~\ref{tab:e2_full} reports the full token-budget sweep on the $N=338$ post-cutoff arXiv test split across strict summary budgets $K \in \{100, 300, 500, 800\}$, alongside unconstrained Prompt-Only reference runs at $K \in \{1{,}200, 2{,}000\}$ (plotted in Figure~\ref{fig:scaling}(a)). MaxIM shifts the entire compression frontier upward: at $K=100$ ($95.8$ mean tokens), MaxIM reaches $34.26\%$ Abstain-MCQ accuracy, outperforming Prompt-Only at $K=500$ ($31.30\%$, $469.0$ tokens; $\mathbf{4.9\times}$ \textbf{fewer tokens}). At $K=500$ ($66.39\%$, $427.7$ tokens), MaxIM surpasses Prompt-Only even at $K=2{,}000$ ($64.32\%$, $1{,}453.3$ tokens; $\mathbf{3.4\times}$ \textbf{compression gain}) before saturating at $67.04\%$ ($K=800$), consistent with Proposition~\ref{prop:capacity_transfer}. Notably, if abstention (option E) is removed, standard 4-way Forced-MCQ accuracy compresses into a narrow $87.7\%$--$94.0\%$ band across all methods and budgets (e.g., Prompt-Only already scores $88.28\%$ at $K=100$ vs.\ $92.07\%$ for MaxIM at $K=500$): once any coarse summary identifies a paper's topic, the frozen solver eliminates implausible distractors via parametric knowledge, confirming that explicit abstention is essential to measure evidence actually retained in $z_T$.

Table~\ref{tab:fig2b_incorp} details the 5-chunk-window claim incorporation rates shown in Figure~\ref{fig:scaling}(b). During chunks $1$--$5$, spare buffer capacity allows all methods to incorporate $44.9\%$--$56.5\%$ of incoming chunk claims. Once the $K=500$ budget fills (chunks $11$--$25$), unguided Prompt-Only drops to $15.4\%$ because it avoids overwriting prior text. MI-Only RL ($23.3\%$ in chunks $21$--$25$) packs claims aggressively from the start ($56.5\%$) under consistency shaping, whereas full MaxIM sustains the highest late-paper incorporation ($\mathbf{26.8\%}$ in chunks $21$--$25$) by selectively evicting low-utility background prose to retain late methodological and experimental findings.

\begin{table}[htbp]
\centering
\caption{\textbf{Full arXiv token-budget Pareto scaling ($N=338$).} Average $\pm$ SEM across episodes (SentencePiece token counts under Gemini 2.5 Flash tokenizer).}
\label{tab:e2_full}
\vspace{2pt}
{\footnotesize
\medmuskip=1mu
\setlength{\tabcolsep}{4.0pt}
\begin{tabular}{llcccc}
\toprule
\textbf{Budget} & \textbf{Method} & \mcell{\textbf{Abstain-}\\\textbf{MCQ $\uparrow$}} & \mcell{\textbf{Abstain}\\\textbf{\% $\downarrow$}} & \mcell{\textbf{Avg.}\\\textbf{Tokens}} & \mcell{\textbf{Terminal}\\\textbf{Tokens}} \\
\midrule
arXiv $K{=}100$ & Prompt-Only & $7.04 \pm 0.69$ & $92.07 \pm 0.70$ & $97.7 \pm 0.1$ & $98.1 \pm 0.2$ \\
arXiv $K{=}100$ & \textbf{MaxIM (Ours)} & $\mathbf{34.26 \pm 1.27}$ & $\mathbf{64.32 \pm 1.28}$ & $\mathbf{95.8 \pm 0.1}$ & $98.7 \pm 0.1$ \\
\midrule
arXiv $K{=}300$ & Prompt-Only & $22.43 \pm 1.11$ & $76.75 \pm 1.13$ & $284.2 \pm 0.6$ & $297.7 \pm 1.2$ \\
arXiv $K{=}300$ & \textbf{MaxIM (Ours)} & $\mathbf{50.06 \pm 1.35}$ & $\mathbf{48.40 \pm 1.36}$ & $\mathbf{276.8 \pm 0.6}$ & $298.3 \pm 0.4$ \\
\midrule
arXiv $K{=}500$ & Prompt-Only & $31.30 \pm 1.28$ & $68.11 \pm 1.29$ & $469.0 \pm 0.8$ & $500.3 \pm 1.6$ \\
arXiv $K{=}500$ & \textbf{MaxIM (Ours)} & $\mathbf{66.39 \pm 1.31}$ & $\mathbf{32.01 \pm 1.31}$ & $\mathbf{427.7 \pm 1.5}$ & $\mathbf{496.7 \pm 0.5}$ \\
\midrule
arXiv $K{=}800$ & Prompt-Only & $38.88 \pm 1.34$ & $60.30 \pm 1.35$ & $730.0 \pm 1.4$ & $800.6 \pm 2.1$ \\
arXiv $K{=}800$ & \textbf{MaxIM (Ours)} & $\mathbf{67.04 \pm 1.30}$ & $\mathbf{31.60 \pm 1.30}$ & $\mathbf{653.9 \pm 3.0}$ & $\mathbf{789.6 \pm 1.8}$ \\
\midrule
arXiv $1.2\text{K}/2\text{K}$ & Prompt-Only (Ref.) & $49.35/64.32$ & $49.76/34.73$ & $1036.9/1453.3$ & $1195.0/1899.9$ \\
\bottomrule
\end{tabular}%
}
\end{table}
\begin{table}[htbp]
\centering
\caption{\textbf{arXiv turn-wise chunk claim incorporation (\%) across 5-chunk streaming windows ($N=338$, $K=500$; supporting Figure~\ref{fig:scaling}(b)).} Entries measure the percentage of atomic claims in incoming chunk $o_t$ (extracted by Gemini 2.5 Flash-Lite) entailed by updated summary $z_t$ (verified by Gemini 3.0 Flash).}
\label{tab:fig2b_incorp}
\vspace{2pt}
{\footnotesize
\setlength{\tabcolsep}{5.2pt}
\begin{tabular}{lccccc}
\toprule
\textbf{Method} & \textbf{Chunks $1$--$5$} & \textbf{Chunks $6$--$10$} & \textbf{Chunks $11$--$15$} & \textbf{Chunks $16$--$20$} & \textbf{Chunks $21$--$25$} \\
\midrule
Prompt-Only & $49.8\%$ & $31.7\%$ & $19.5\%$ & $15.9\%$ & $15.4\%$ \\
Policy SFT & $48.5\%$ & $31.4\%$ & $22.6\%$ & $18.0\%$ & $17.6\%$ \\
MemAgent RL ($\gamma{=}1.0$) & $44.9\%$ & $38.3\%$ & $25.7\%$ & $19.2\%$ & $18.6\%$ \\
Task-Only RL ($\gamma{=}0.9$) & $50.2\%$ & $40.0\%$ & $27.0\%$ & $21.0\%$ & $20.4\%$ \\
MI-Only RL ($R_{\text{task}}{=}0$) & $\mathbf{56.5\%}$ & $\mathbf{44.2\%}$ & \underline{$30.5\%$} & \underline{$24.0\%$} & \underline{$23.3\%$} \\
\textbf{MaxIM (Full, Ours)} & \underline{$50.2\%$} & $\mathbf{44.2\%}$ & $\mathbf{33.9\%}$ & $\mathbf{27.3\%}$ & $\mathbf{26.8\%}$ \\
\bottomrule
\end{tabular}%
}
\end{table}

\subsection{Offline Critic Validation: Raw VPL vs.\ PMI}
\label{app:critic_probes}

Table~\ref{tab:critic_probes} compares unnormalized conditional log-likelihood (raw VPL) against our baseline-subtracted PMI critic $q_\phi$ (Section~\ref{subsec:objective}) on held-out trajectories and synthetic quality triplets ($z_t^{\text{hq}}, z_t^{\text{vague}}, z_t^{\text{halluc}}$) prior to RL. Raw sufficiency log-likelihood $\log q_\phi(Y_t \mid z_t)$ is dominated by cross-paper question difficulty, yielding tiny within-paper spread ($\sigma_{\text{within}} = 0.043$ nats) and near-zero correlation with downstream Abstain-MCQ utility ($r = +0.050$). Subtracting the masked-summary baseline $\log q_\phi(Y_t \mid z_{\mathrm{null}})$ cancels paper difficulty, expanding within-paper spread by $8.1\times$ ($0.351$ nats) and raising utility correlation to $r = +0.324$ (with $100\%$ pairwise accuracy separating HQ from vague or hallucinated WildChat belief states). Similarly, forward consistency PMI relative to $\tilde{H}_{\mathrm{null}}$ removes raw likelihood's preference for longer summaries ($r(\text{tokens})$ shifts from $+0.323$ to $-0.170$) and boosts HQ-vs.-hallucinated classification AUROC from $0.685$ to $\mathbf{0.963}$.

\begin{table}[htbp]
\centering
\caption{\textbf{Offline critic validation comparing Raw VPL vs.\ PMI} on held-out arXiv trajectories/triplets and held-out WildChat validation triplets.}
\label{tab:critic_probes}
\vspace{2pt}
{\footnotesize
\setlength{\tabcolsep}{4.0pt}
\begin{tabular}{lcc}
\toprule
\textbf{Validation Metric} & \mcell{\textbf{Raw VPL}\\\textbf{Critic}} & \mcell{\textbf{PMI Critic}\\\textbf{(Ours)}} \\
\midrule
arXiv Within-Paper Sufficiency Spread ($\sigma_{\text{within}}$, nats $\uparrow$) & $0.043$ & $\mathbf{0.351}$ ($8.1\times$) \\
arXiv Correlation with Downstream Abstain-MCQ ($r \uparrow$) & $+0.050$ & $\mathbf{+0.324}$ ($6.5\times$) \\
arXiv Hallucination Discrimination AUROC (HQ vs.\ Halluc.\ $\uparrow$) & $0.685$ & $\mathbf{0.963}$ ($+27.8$ pts) \\
arXiv Grounding Length Correlation ($r(\text{tokens})$, closer to $0$) & $+0.323$ & $\mathbf{-0.170}$ \\
WildChat Sufficiency Pairwise Acc.\ (HQ $>$ Vague / Halluc.) & --- & $\mathbf{100.0\%}$ / $\mathbf{100.0\%}$ \\
WildChat Recursive Consistency Pairwise Acc.\ (HQ $>$ Halluc.) & --- & $\mathbf{83.6\%}$ ($+0.369$ nats) \\
\bottomrule
\end{tabular}%
}
\end{table}

\subsection{WildChat Downstream Agent Evaluation and Causal Belief-Tracking Fidelity}
\label{app:wc_belief_fidelity}

Table~\ref{tab:wc_downstream_full} expands Table~\ref{tab:main_rq1} with absolute $1$--$10$ GPT-4.1 scores, Coherence ratings, and average conversation lengths across all $N=500$ test dialogues and the fixed $T \ge 10$ cohort ($N=158$). On all $N=500$ seeds, MaxIM ranks first on all four dimensions while resolving user goals in the fewest turns ($8.49$ vs.\ $8.75$ for No-Belief). In paired bootstrap tests, MaxIM significantly outperforms both Task-Only RL ($+0.29$ to $+0.35$ across axes, $p \le 0.0056$) and MemAgent RL ($+0.30$ to $+0.34$, $p \le 0.0064$), as well as Suff-Only RL on Satisfaction ($+0.23$, $p=0.0480$) and Preference Retention ($+0.28$, $p=0.0096$). On the $T \ge 10$ cohort, No-Belief drops by $0.46$--$0.51$ points as early user constraints fall outside the $k=1$ window; here, only MI-shaped policies recover positive gains across every axis, with MaxIM outperforming Task-Only RL by $+0.41$ to $+0.48$ points ($p < 0.05$) and shortening dialogues from $12.61$ to $10.68$ turns.

\begin{table}[htbp]
\centering
\caption{\textbf{Full WildChat bounded-memory ($k=1$, $T \le 15$) downstream chat-agent evaluation} (GPT-4.1 judge without summary access, $1$--$10$ scale). Average $\pm$ SEM (with paired $\Delta$ vs.\ No-Belief in parentheses, computed prior to rounding) on all $N=500$ test seeds and the fixed No-Belief $T \ge 10$ long-horizon cohort ($N=158$).}
\label{tab:wc_downstream_full}
\vspace{2pt}
{\footnotesize
\medmuskip=0mu
\setlength{\tabcolsep}{2.2pt}
\begin{tabular}{lccccc}
\toprule
\textbf{Method} & \textbf{User Sat.\ $\uparrow$} & \textbf{Goal Comp.\ $\uparrow$} & \textbf{Pref.\ Ret.\ $\uparrow$} & \textbf{Coherence $\uparrow$} & \mcell{\textbf{Avg.}\\\textbf{Turns}} \\
\midrule
\multicolumn{6}{l}{\textbf{(i) All Test Dialogues ($N{=}500$)}} \\
\midrule
No-Belief ($k{=}1$) & $8.58 \pm 0.11$ & $8.75 \pm 0.10$ & $8.74 \pm 0.10$ & $8.70 \pm 0.09$ & $8.75$ \\
Prompt-Only & $8.49 \pm 0.11$\,($-0.09$) & $8.66 \pm 0.10$\,($-0.09$) & $8.75 \pm 0.09$\,($+0.01$) & $8.62 \pm 0.10$\,($-0.08$) & $8.74$ \\
Policy SFT & $8.54 \pm 0.11$\,($-0.04$) & $8.70 \pm 0.10$\,($-0.05$) & $8.79 \pm 0.09$\,($+0.05$) & $8.72 \pm 0.09$\,($+0.02$) & $8.61$ \\
MemAgent~\cite{yu2025memagent} & $8.30 \pm 0.11$\,($-0.28$) & $8.48 \pm 0.10$\,($-0.27$) & $8.60 \pm 0.10$\,($-0.13$) & $8.44 \pm 0.11$\,($-0.26$) & $8.84$ \\
Task-Only RL & $8.28 \pm 0.12$\,($-0.30$) & $8.46 \pm 0.11$\,($-0.29$) & $8.58 \pm 0.10$\,($-0.15$) & $8.47 \pm 0.10$\,($-0.23$) & $8.58$ \\
MI-Only RL & \underline{$8.58 \pm 0.10$\,($0.00$)} & \underline{$8.74 \pm 0.09$\,($0.00$)} & \underline{$8.81 \pm 0.09$\,($+0.07$)} & \underline{$8.73 \pm 0.09$\,($+0.03$)} & $8.71$ \\
Suff-Only RL & $8.40 \pm 0.11$\,($-0.18$) & $8.60 \pm 0.10$\,($-0.14$) & $8.62 \pm 0.10$\,($-0.12$) & $8.56 \pm 0.10$\,($-0.14$) & $8.57$ \\
\textbf{MaxIM (Full)} & $\mathbf{8.63 \pm 0.10}$\,($\mathbf{+0.05}$) & $\mathbf{8.80 \pm 0.09}$\,($\mathbf{+0.05}$) & $\mathbf{8.90 \pm 0.08}$\,($\mathbf{+0.16}$) & $\mathbf{8.77 \pm 0.09}$\,($\mathbf{+0.06}$) & $\mathbf{8.49}$ \\
\midrule
\multicolumn{6}{l}{\textbf{(ii) Fixed Long-Horizon Cohort ($T{\ge}10$, $N{=}158$)}} \\
\midrule
No-Belief ($k{=}1$) & $8.08 \pm 0.20$ & $8.36 \pm 0.18$ & $8.28 \pm 0.19$ & $8.19 \pm 0.18$ & $12.61$ \\
Prompt-Only & $7.99 \pm 0.20$\,($-0.09$) & $8.23 \pm 0.19$\,($-0.13$) & $8.35 \pm 0.19$\,($+0.08$) & $8.15 \pm 0.19$\,($-0.04$) & $11.12$ \\
Policy SFT & $7.87 \pm 0.20$\,($-0.21$) & $8.15 \pm 0.18$\,($-0.21$) & $8.28 \pm 0.18$\,($0.00$) & $8.14 \pm 0.18$\,($-0.05$) & $11.08$ \\
MemAgent~\cite{yu2025memagent} & $7.94 \pm 0.21$\,($-0.15$) & $8.15 \pm 0.20$\,($-0.21$) & $8.30 \pm 0.19$\,($+0.02$) & $8.06 \pm 0.20$\,($-0.13$) & $11.12$ \\
Task-Only RL & $7.72 \pm 0.22$\,($-0.37$) & $7.98 \pm 0.20$\,($-0.38$) & $8.17 \pm 0.19$\,($-0.11$) & $7.90 \pm 0.20$\,($-0.29$) & $10.84$ \\
MI-Only RL & $\mathbf{8.26 \pm 0.19}$\,($\mathbf{+0.18}$) & \underline{$8.48 \pm 0.18$\,($+0.12$)} & \underline{$8.51 \pm 0.17$\,($+0.23$)} & $\mathbf{8.44 \pm 0.17}$\,($\mathbf{+0.25}$) & $10.91$ \\
Suff-Only RL & \underline{$8.25 \pm 0.20$\,($+0.16$)} & $\mathbf{8.49 \pm 0.18}$\,($\mathbf{+0.13}$) & \underline{$8.51 \pm 0.19$\,($+0.23$)} & \underline{$8.36 \pm 0.18$\,($+0.17$)} & $10.89$ \\
\textbf{MaxIM (Full)} & $8.20 \pm 0.19$\,($+0.11$) & $8.42 \pm 0.17$\,($+0.06$) & $\mathbf{8.58 \pm 0.16}$\,($\mathbf{+0.30}$) & $8.32 \pm 0.17$\,($+0.13$) & $\mathbf{10.68}$ \\
\bottomrule
\end{tabular}%
}
\end{table}

Table~\ref{tab:wc_belief_fidelity} evaluates the $1$--$10$ causal belief-tracking reward $R_{\text{task}}$ directly on held-out Test ($N=500$), unseen-simulator OOD ($N=200$), and $T=15$ horizon extrapolation. Together, Tables~\ref{tab:wc_downstream_full} and~\ref{tab:wc_belief_fidelity} expose a striking reward-to-utility gap: MemAgent RL ($8.98$) and Task-Only RL ($8.79$) nearly match MaxIM ($\mathbf{9.00}$ Test; $\mathbf{9.05}$ OOD) on terminal belief accuracy, yet both degrade live assistant performance below No-Belief. Because sparse terminal $R_{\text{task}}$ evaluates only the final state, unshaped RL tolerates intermediate-turn confabulations ($7.04\%$--$9.35\%$ contradictions vs.\ MaxIM's $4.27\%$; Table~\ref{tab:grounding_rq2}) and suffers sharper horizon decay at $T=15$ ($7.79$ and $8.18$ vs.\ MaxIM's $\mathbf{8.50}$), misleading the assistant on earlier turns.

\begin{table}[htbp]
\centering
\caption{\textbf{WildChat $1$--$10$ causal belief-state tracking fidelity} (training reward metric) across held-out Test ($N=500$), unseen-simulator OOD ($N=200$), and $T=15$ horizon extrapolation.}
\label{tab:wc_belief_fidelity}
\vspace{2pt}
{\small
\begin{tabular}{lccc}
\toprule
\textbf{Method} & \textbf{Test ($N{=}500$)} & \textbf{Unseen-Sim OOD ($N{=}200$)} & \textbf{Turn $15$ Extrap.} \\
\midrule
Prompt-Only & $8.32 \pm 0.07$ & $8.44 \pm 0.11$ & $7.11$ \\
Policy SFT & $8.31 \pm 0.06$ & $8.29 \pm 0.09$ & $6.96$ \\
MI-Only RL ($R_{\text{task}}{=}0$) & $8.33 \pm 0.07$ & $8.32 \pm 0.10$ & $6.85$ \\
Suff-Only RL ($w_c{=}0$) & $8.76 \pm 0.06$ & $8.76 \pm 0.09$ & $7.50$ \\
Task-Only RL ($\gamma{=}0.9$) & $8.79 \pm 0.06$ & $8.90 \pm 0.08$ & $7.79$ \\
MemAgent RL ($\gamma{=}1.0$)~\cite{yu2025memagent} & \underline{$8.98 \pm 0.05$} & \underline{$8.92 \pm 0.09$} & \underline{$8.18$} \\
\textbf{MaxIM (Full, Ours)} & $\mathbf{9.00 \pm 0.05}$ & $\mathbf{9.05 \pm 0.08}$ & $\mathbf{8.50}$ \\
\bottomrule
\end{tabular}%
}
\end{table}

\subsection{Turn-Stratified Qualitative and Cross-Judge Grounding Evaluation}
\label{app:turnwise_e4}

\paragraph{Cross-Family Grounding Verification.}
To verify that Table~\ref{tab:grounding_rq2}'s arXiv grounding results are not specific to GPT-4.1, we re-evaluate terminal-$z_T$ atomic-claim entailment using \textbf{Gemini 3.0 Flash}. We observe the exact same pattern: all RL policies maintain high factual precision---$98.43\%$ (Suff-Only RL), $98.19\%$ (MemAgent RL), $98.14\%$ (Task-Only RL), $97.77\%$ (MI-Only RL), and $96.76\%$ (MaxIM), alongside $98.44\%$ for Prompt-Only---whereas Policy SFT drops to $74.70\%$ as imitation errors compound over $25$ updates.

\paragraph{Turn-Stratified Qualitative Dynamics.}
Table~\ref{tab:turnwise_e4} tracks GPT-4.1 qualitative ratings across intermediate and terminal turns on both benchmarks. On arXiv (Table~\ref{tab:turnwise_e4}(a)), Policy SFT achieves high Usefulness at $t=3$ ($77.0\%$) while summaries remain short, but collapses to $13.4\%$ by $t=20$ as ungrounded claims accumulate. By contrast, MaxIM climbs from $39.0\%$ ($t=3$) to $78.0\%$ ($t=15$) and $72.9\%$ ($t=20$) against Prompt-Only while achieving the highest standalone Consistency score ($7.55$ vs.\ $5.99$--$6.44$ across other RL methods). Furthermore, when GPT-4.1 compares MaxIM directly against MemAgent on the exact same paper prefixes, MaxIM wins on Usefulness at every evaluated chunk ($72.0\%$ at $t=3$ through $54.8\%$ at $t=20$, plus a $61.6\%$ Consistency win rate at $t=20$).

On WildChat (Table~\ref{tab:turnwise_e4}(b)), standalone $1$--$10$ Consistency saturates near ceiling across all RL policies ($9.69$--$9.86$ vs.\ $9.64$ for Prompt-Only and $9.37$ for Policy SFT; MaxIM scores $9.73 \pm 0.04$) because unchanged fields of the 10-key JSON state carry forward smoothly across turns. Pairwise Usefulness and Truthfulness reveal a clear horizon crossover: at early turns ($t=3, 5$), Task-Only RL's speculative trait guesses slightly edge MaxIM on Usefulness ($82.1\%$ vs.\ $79.5\%$ at $t=5$) at the cost of lower Truthfulness ($70.5\%$ vs.\ $\mathbf{78.6\%}$). Once conversations extend to $t \ge 8$, MaxIM leads every baseline on \emph{both} Usefulness ($76.4\%$--$90.2\%$) and Truthfulness ($71.5\%$--$88.5\%$).

\begin{table}[htbp]
\centering
\caption{\textbf{Turn-stratified GPT-4.1 qualitative evaluation.} \textbf{(a)} arXiv pairwise Usefulness win rate (\%) vs.\ Prompt-Only across chunks $t$ (plus standalone $1$--$10$ Consistency at $t=20$, and direct MaxIM vs.\ MemAgent judging). \textbf{(b)} WildChat pairwise Usefulness / Truthfulness win rates (\%) vs.\ Prompt-Only across turns (active conversation counts $N$ shown per column on a $100$-seed replay subset alongside all $N=500$ terminal states).}
\label{tab:turnwise_e4}
\vspace{2pt}
{\footnotesize
\setlength{\tabcolsep}{3.0pt}
\begin{tabular}{lccccccc}
\toprule
\multicolumn{8}{l}{\textbf{(a) arXiv Pairwise Summary Comparison ($N{=}338$; vs.\ Prompt-Only unless noted)}} \\
\midrule
\textbf{Method} & $t = 3$ & $t = 5$ & $t = 10$ & $t = 15$ & \multicolumn{3}{c}{$t = 20$ (\textbf{Use \% / Cons.})} \\
\midrule
Policy SFT & $77.0\%$ & $35.0\%$ & $26.0\%$ & $19.0\%$ & \multicolumn{3}{c}{$13.4\%$ / $7.12$} \\
MI-Only RL & $40.0\%$ & $38.0\%$ & $47.0\%$ & $47.0\%$ & \multicolumn{3}{c}{$47.0\%$ / $6.04$} \\
Task-Only RL & $42.0\%$ & $52.0\%$ & $61.0\%$ & $69.0\%$ & \multicolumn{3}{c}{$62.5\%$ / $6.44$} \\
Suff-Only RL & $72.0\%$ & $78.0\%$ & $80.0\%$ & $81.0\%$ & \multicolumn{3}{c}{$74.4\%$ / $5.99$} \\
MemAgent RL & $31.0\%$ & $48.0\%$ & $74.0\%$ & $87.0\%$ & \multicolumn{3}{c}{$80.1\%$ / $6.07$} \\
\textbf{MaxIM (Full)} & $39.0\%$ & $50.0\%$ & $72.0\%$ & $78.0\%$ & \multicolumn{3}{c}{$72.9\%$ / $\mathbf{7.55}$} \\
\textbf{MaxIM} vs.\ \textbf{MemAgent} & $\mathbf{72.0\%}$ & $\mathbf{62.0\%}$ & $\mathbf{57.0\%}$ & $\mathbf{56.0\%}$ & \multicolumn{3}{c}{$\mathbf{54.8\%}$ ($\mathbf{61.6\%}$ Cons.\ win)} \\
\midrule
\multicolumn{8}{l}{\textbf{(b) WildChat Belief-State Usefulness / Truthfulness Win \% vs.\ Prompt-Only}} \\
\midrule
\textbf{Method} & \mcell{\textbf{Term.}\\{\scriptsize($N{=}500$)}} & \mcell{$t = 3$\\{\scriptsize($N{=}99$)}} & \mcell{$t = 5$\\{\scriptsize($N{=}96$)}} & \mcell{$t = 8$\\{\scriptsize($N{=}65$)}} & \mcell{$t = 10$\\{\scriptsize($N{=}37$)}} & \mcell{$t = 12$\\{\scriptsize($N{=}21$)}} & \mcell{$t = 15$\\{\scriptsize($N{=}12$)}} \\
\midrule
Policy SFT & $37.6/36.2$ & $50.0/44.1$ & $40.2/40.2$ & $38.2/37.4$ & $37.7/36.1$ & $40.6/34.4$ & $51.6/51.6$ \\
MI-Only RL & $40.8/40.8$ & $49.0/52.0$ & $42.9/43.8$ & $42.3/45.5$ & $37.7/41.0$ & $43.8/37.5$ & $51.6/50.0$ \\
Task-Only RL & $\mathbf{79.8}/69.8$ & $\mathbf{78.4}/64.7$ & $\mathbf{82.1}/70.5$ & $74.8/67.5$ & $78.7/75.4$ & $78.1/75.0$ & $79.0/71.0$ \\
Suff-Only RL & $72.2/65.8$ & $67.6/62.7$ & $69.6/61.6$ & $73.2/63.4$ & $68.9/60.7$ & $81.2/75.0$ & $69.4/61.3$ \\
MemAgent RL & $66.4/61.0$ & $67.6/57.8$ & $67.0/61.6$ & $70.7/63.4$ & $73.8/65.6$ & $75.0/62.5$ & $61.3/58.1$ \\
\textbf{MaxIM (Full)} & \underline{$73.8$}/$\mathbf{72.4}$ & \underline{$72.5$}/$\mathbf{65.7}$ & \underline{$79.5$}/$\mathbf{78.6}$ & $\mathbf{76.4}/\mathbf{71.5}$ & $\mathbf{90.2}/\mathbf{88.5}$ & $\mathbf{87.5}/\mathbf{78.1}$ & $\mathbf{80.6}/\mathbf{75.8}$ \\
\bottomrule
\end{tabular}%
}
\end{table}

\clearpage

\section{Broader Impacts and Limitations}
\label{app:broader-impact-and-limitations}

\subsection{Broader Impacts}
\label{app:broader-impacts}

MaxIM introduces a principled approach to bounded memory management in streaming environments. On the positive side, this framework enhances the reliability of long-horizon AI agents by actively suppressing hallucinations and enforcing factual grounding through recursive consistency. This is particularly beneficial for high-stakes applications requiring persistent context---such as educational tutoring, accessibility tools, and medical or legal document analysis---where unbounded context scaling is computationally prohibitive.

However, the capability to persistently track dynamically evolving latent states (such as user emotions or implicit intents, as demonstrated in our WildChat experiments) poses potential privacy and surveillance risks. If deployed without safeguards, agents equipped with highly effective intent-tracking memory could be used for manipulative profiling or invasive monitoring. Furthermore, because the memory update relies heavily on its own prior state, early biases or misinterpretations encoded in the summary may compound over time. Mitigating these risks requires transparent user consent for long-term memory tracking, routine auditing of the internal summary states for bias or sensitive personal information, and the implementation of robust safety filters on the downstream action agent.

\subsection{Limitations}
\label{app:limitations}

While MaxIM effectively mitigates credit assignment failures through information-theoretic shaping, we acknowledge several limitations that provide avenues for future work:
\begin{itemize}[leftmargin=1.4em, itemsep=3pt]
    \item \textbf{The Markovian approximation and irreversible loss:} MaxIM enforces a strict Markovian bottleneck by conditioning solely on the immediate local context $\tilde{H}_t = (z_{t-1}, a_{t-1}, o_t)$. During inference, if the policy erroneously discards a critical piece of information at step $t$, the loss is irreversible because the model lacks attention mechanisms to re-access $H_{t-2}$. Pairing MaxIM's compressed working memory with an external episodic retrieval fallback is a promising hybrid architecture for deployment.
    \item \textbf{Oracle dependency for critic training:} MaxIM circumvents the need for expensive human preference labels, shifting supervision to an offline synthetic pipeline (Appendix~\ref{app:data_generation}) where privileged target variables $Y_t$ are available strictly during training. Applying MaxIM to a new domain requires defining a task-relevant target proxy $Y_t$ and generating offline critic data; as quantified in Proposition~\ref{prop:action_dependent} of Appendix~\ref{app:proof_thm1}, misspecifying $Y_t$ introduces an additive mediation floor proportional to $(\varepsilon_r, \varepsilon_T)$.
    \item \textbf{Training compute overhead:} While step-wise teacher-forced PMI rewards eliminate autoregressive critic generation and incur zero test-time overhead (Table~\ref{tab:compute_cost}), the RL fine-tuning phase requires critic forward passes during the shaping window ($s < 600$), adding training compute relative to sparse-reward RL.
    \item \textbf{LLM-as-a-Judge evaluation bias:} Using a single model family for both policy training and qualitative evaluation can introduce self-preference bias. To mitigate this risk, we conduct atomic-claim source grounding and qualitative evaluations using an independent cross-family judge (\textbf{GPT-4.1}) alongside Gemini 3.0 Flash; nonetheless, live human-in-the-loop user studies would further complement simulated multi-turn benchmarks.
\end{itemize}

\end{document}